\PassOptionsToPackage{table}{xcolor}
\pdfoutput=1
\documentclass[11pt]{article}
\usepackage[preprint]{acl}
\usepackage{times}
\usepackage{latexsym}
\usepackage[T1]{fontenc}
\usepackage[utf8]{inputenc}
\usepackage{microtype}
\usepackage{inconsolata}
\usepackage{amsmath}
\usepackage{booktabs}
\usepackage{multirow}
\usepackage{amssymb}
\usepackage{balance}
\usepackage{subcaption}

\usepackage{pifont}
\usepackage{makecell}
\usepackage{filecontents}
\usepackage{bbm}
\usepackage{pgfplots}
\usetikzlibrary{patterns}
\usepackage[inline,shortlabels]{enumitem}
\usepackage{natbib}
\usepackage{csquotes}
\usepackage{hyperref}
\usepackage{url}
\usepackage{CJKutf8}
\usepackage{algorithm, algorithmicx, algpseudocode}
\usepackage{svg}
\usepackage{bm}
\usepackage{graphics}
\usepackage{graphicx}
\usepackage{array}
\usepackage[english]{babel}
\usepackage{textcomp}
\usepackage{latexsym}
\usepackage{siunitx}  
\usepackage[normalem]{ulem}
\useunder{\uline}{\ul}{}
\usepackage{array}
\usepackage{titletoc}
\usepackage{tikz}
\usetikzlibrary{tikzmark}
\makeatletter
\newcommand*\myfontsize{%
  \@setfontsize\myfontsize{7}{8}%
}
\makeatother
\newcommand{\mytextbox}[2]{\tikzmarknode[draw=#1,thick,inner sep=2pt]{test}{\myfontsize #2}}
\definecolor{myred}{rgb}{0.7, 0.3, 0.0}
\definecolor{myblue}{rgb}{0.2, 0.3, 0.6}
\definecolor{mygreen}{rgb}{0, 0.6, 0.3}

\newcommand{\red}[1]{\mytextbox{myred}{\textbf{\textcolor{myred}{#1}}}}
\newcommand{\blue}[1]{\mytextbox{myblue}{\textbf{\textcolor{myblue}{#1}}}}

\title{RetroLLM: Empowering Large Language Models to Retrieve Fine-grained Evidence within Generation}

\author{%
Xiaoxi Li$^1$, Jiajie Jin$^1$, Yujia Zhou$^2$, Yongkang Wu$^3$, Zhonghua Li$^3$,\\
\textbf{Qi Ye$^3$,} \textbf{Zhicheng Dou$^1$}\thanks{Correpsonding author.} \\
$^1$Gaoling School of Artificial Intelligence, Renmin University of China\\
$^2$Tsinghua University~~$^3$Huawei Poisson Lab\\
\texttt{\{xiaoxi\_li, dou\}@ruc.edu.cn} \\
}

\begin{document}
\begin{CJK}{UTF8}{gbsn}

\maketitle

\begin{abstract}
Large language models (LLMs) exhibit remarkable generative capabilities but often suffer from hallucinations. Retrieval-augmented generation (RAG) offers an effective solution by incorporating external knowledge, but existing methods still face several limitations: additional deployment costs of separate retrievers, redundant input tokens from retrieved text chunks, and the lack of joint optimization of retrieval and generation. To address these issues, we propose \textbf{RetroLLM}, a unified framework that integrates retrieval and generation into a single, cohesive process, enabling LLMs to directly generate fine-grained evidence from the corpus with constrained decoding. 
Moreover, to mitigate false pruning in the process of constrained evidence generation, we introduce (1) hierarchical FM-Index constraints, which generate corpus-constrained clues to identify a subset of relevant documents before evidence generation, reducing irrelevant decoding space; and (2) a forward-looking constrained decoding strategy, which considers the relevance of future sequences to improve evidence accuracy.
Extensive experiments on five open-domain QA datasets demonstrate RetroLLM's superior performance across both in-domain and out-of-domain tasks.
The code is available at \url{https://github.com/sunnynexus/RetroLLM}.

\end{abstract}

\begin{figure*}[!t]
    \centering
    \includegraphics[width=0.93\linewidth]{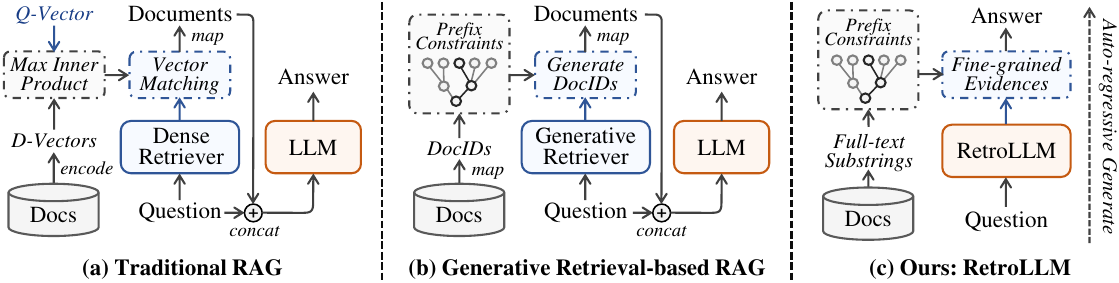}
    \caption{
    Comparison of retrieval-augmented generation frameworks. (a) Traditional RAG uses a dense retriever for document matching, while (b) generative RAG relies on constrained DocID generation. Both require feeding retrieved document text into the LLM for answer generation. (c) Our RetroLLM unifies retrieval and generation in a single auto-regressive decoding process, leveraging FM-Index constraints to retrieve fine-grained evidence.
    }
    \label{fig:compare} 
\end{figure*}

\section{Introduction}

Large language models (LLMs) exhibit remarkable capabilities and are widely applied in various domains~\cite{llm-survey, llm4ir, mo2024_conv_survey}. However, due to their reliance on model internal memory, they often struggle with long-tail or newly updated knowledge, leading to the issue of ``hallucinations''~\cite{survey_hallu_llm}. To address this, retrieval-augmented generation (RAG) has emerged as a promising solution. By integrating retrieval of external knowledge, RAG enables models to access up-to-date and factual information, enhancing the accuracy and reliability of their responses~\cite{rag, ragsurvey}.

Existing RAG methods typically rely on a separate dense retriever to fetch top-$k$ text chunks for LLMs to generate answers, as shown in Figure~\ref{fig:compare}(a). However, these methods face several limitations: (1) Maintaining a separate retriever increases deployment costs~\cite{2409_onegen}; (2) Retrieved documents often contain redundant information, consuming vast input tokens and distracting the model's attention from key information~\cite{llmlingua}; (3) The fixed granularity and number of retrieved text chunks limits flexibility of RAG systems~\cite{24_acl_chunking_free}; and (4) Retrievers rely on standalone document indices, hindering joint optimization with generators. Since retrieval and generation are inherently interconnected, jointly learning these tasks can enhance the overall performance of RAG systems~\cite{rag, unigen}. Thus, we aim to develop a unified framework that seamlessly integrates retrieval and generation processes.

Recently, generative retrieval (GR) has emerged as a promising approach that leverages generative models to directly generate document identifiers (DocIDs), eliminating the need for document indices and making it possible for joint optimization~\cite{genir_survey, dsi, unigen, 2402_corpuslm}. However, existing GR methods still require mapping the generated DocIDs back to the document content before these can be used by LLMs for answer generation, as depicted in Figure~\ref{fig:compare}(b). This step disrupts the seamless integration of retrieval and generation processes. 

To address the above limitations, we propose \textbf{RetroLLM}, which empowers LLMs to generate factual evidence from knowledge sources and final answer within a unified, auto-regressive decoding process, as shown in Figure~\ref{fig:compare}(c). RetroLLM enables the model to autonomously decide how much evidence to retrieve and when to generate the final response, eliminating the need for a separate embedding model and enhancing the flexibility of the RAG system. Furthermore, RetroLLM achieves joint optimization of retrieval and generation, facilitating a deeper understanding of their relationships and improving overall performance.

To achieve this, a simple approach is to apply constrained beam search based on FM-Index to generate factual evidence contained in corpus~\cite{2407_riches}. However, the prefix-constrained approach suffers from severe \textbf{false pruning} problem, where correct evidence sequences are often pruned due to errors in early decoding steps~\cite{autotsg}. While initial prefixes may appear relevant,  subsequent decoded sequences often reveal that they are grounded in irrelevant documents, leading to failure in generating relevant evidence. This issue arises from two main challenges: (1) Large corpora result in a vast number of prefix choices during early constrained decoding steps, making it difficult to choose the correct one; (2) It is also difficult to predict the relevance of subsequent sequences based solely on a short prefix. (See Appendix~\ref{app:false_pruning} for details)

To address these challenges, we propose two key strategies: (1) We construct a \textbf{hierarchical FM-Index}, which first generates corpus-constrained clues to identify a subset of candidate documents. Evidence is then generated under the constraints of this subset's FM-Index, significantly reducing the irrelevant decoding space, especially in the early steps. 
(2) For evidence generation, we introduce \textbf{forward-looking constrained decoding}, which utilizes the document FM-Index to identify future windows within the candidate documents based on clues. A relevance model then scores these windows, promoting the generation of relevant evidence.

We conduct extensive experiments on five open-domain QA datasets, testing both in-domain and out-of-domain tasks, different base LLMs, and different parameter sizes. Our experimental results demonstrate the superior performance of RetroLLM compared to traditional RAG and complex RAG strategies.

The main contributions of this paper are: (1) We propose RetroLLM, a unified framework that unifies retrieval and generation into a single auto-regressive process, eliminating the need for a separate retriever and enabling joint optimization of RAG tasks. (2) To reduce irrelevant decoding paths in constrained evidence generation, we propose hierarchical FM-Index constraints and first predict clues with corpus FM-Index constraints to identify a document subset. (3) We introduce forward-looking constrained decoding, which identifies candidate windows based on clues and leverages future window relevance scores to guide the model in generating relevant evidence.

\section{Preliminary}
\subsection{Task Formulation}
Retrieval-augmented generation (RAG) leverages external knowledge sources to enhance the accuracy of language model generations. In this work, we formulate RAG within a generative framework. We divide the task into two sub-problems:

\textbf{Constrained Evidence Generation:} This involves retrieving relevant evidence from a large corpus in a generative manner with pre-built constraints. Formally, let $\mathcal{C}$ be the corpus of documents and $q$ be the input query. The constrained evidence generation process can be formulated as:
\begin{equation}
P(e|q,\mathcal{C}) = \prod\nolimits_{t=1}^{T_e} P(e_t|e_{<t},q,\mathcal{I}_c),
\end{equation}
where $e$ is the generated evidence sequence with length $T_e$, $e_{<t}$ is the tokens generated before position $t$, and $\mathcal{I}_c = \text{FM-Index}(\mathcal{C})$ represents the FM-Index built on corpus $\mathcal{C}$.

\textbf{Answer Generation:} Based on the retrieved evidence, the language model continues to generate the final answer, which can be expressed as:
\begin{equation}
P(a|q,e) = \prod\nolimits_{t=1}^{T_a} P(a_t|a_{<t},q,e),
\end{equation}
where $a$ is the generated answer with length $T_a$ and $a_{<t}$ denotes the tokens generated before position $t$.

\begin{figure}[!t]
\begin{subfigure}{.495\linewidth}
    \centering
    \includegraphics[width=\linewidth]{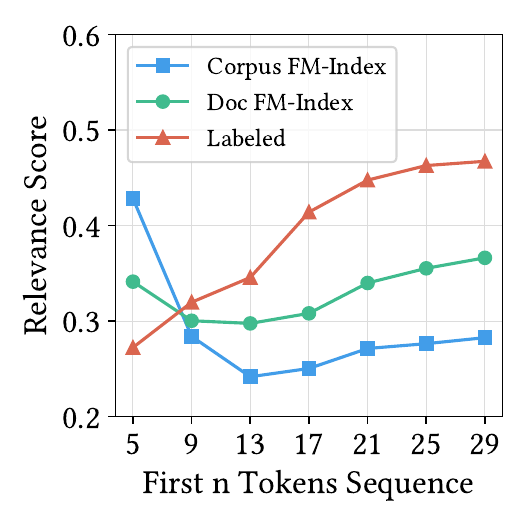}
    \caption{Sequence Relevance}
\end{subfigure}
\begin{subfigure}{.495\linewidth}
    \centering
    \includegraphics[width=\linewidth]{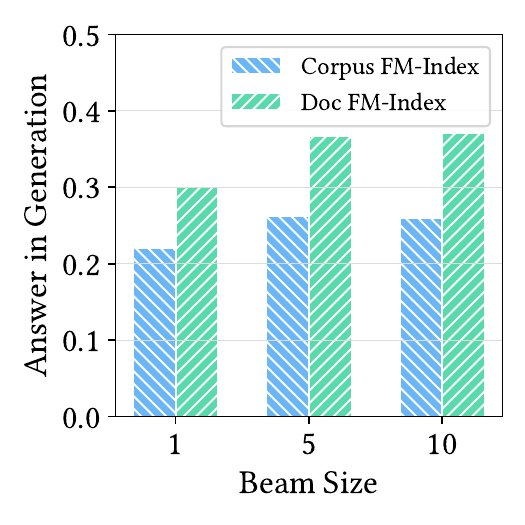}
    \caption{Overall Accuracy}
\end{subfigure}
\caption{Empirical Study on false pruning problem in constrained evidence generation, comparing corpus-level and document-level FM-Index approaches.}
\label{fig:empirical_study}
\end{figure}

\begin{figure*}[!t]
    \centering
    \includegraphics[width=1\linewidth]{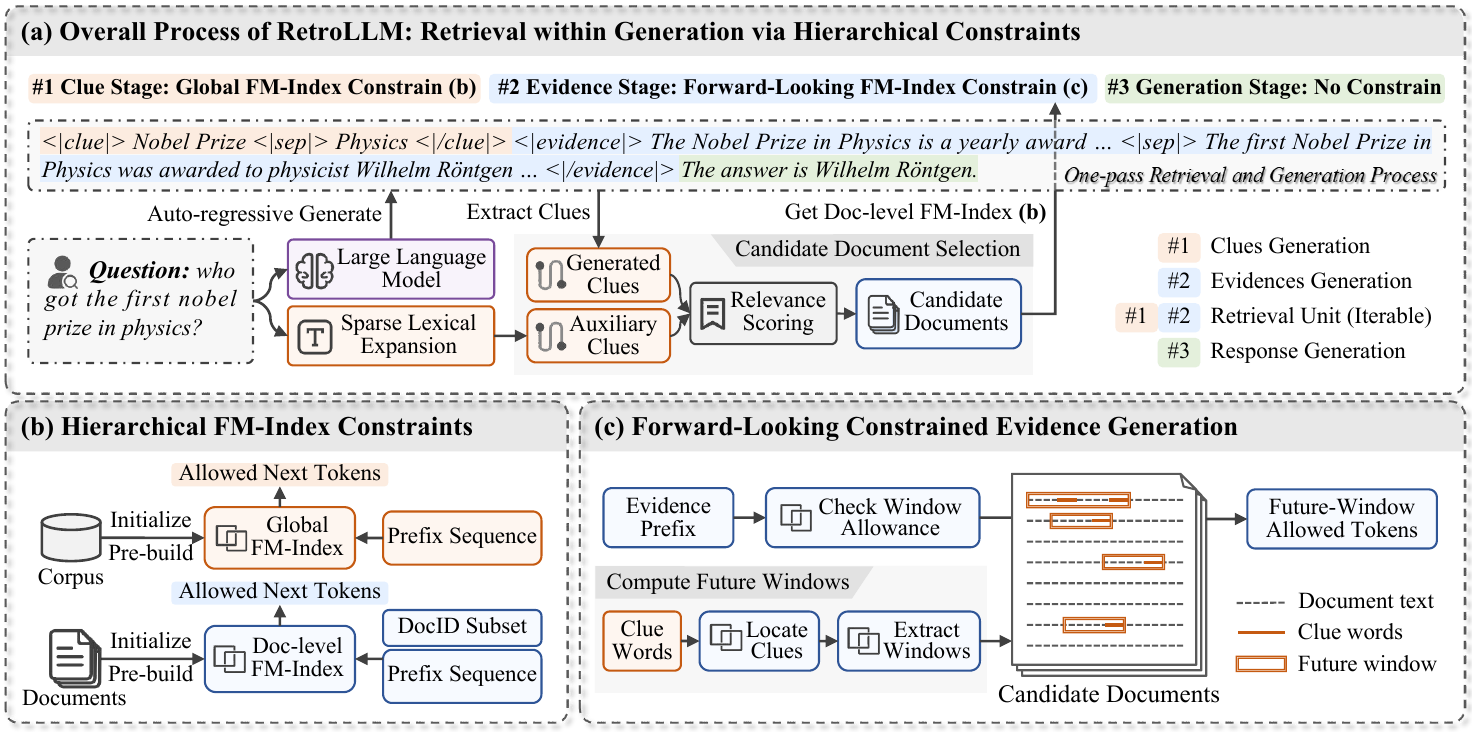}
    \caption{
    Overview of the RetroLLM Framework, which retrieves fine-grained evidence through a hierarchical, forward-looking FM-Index constrained generation process. During generation, the model autonomously determines whether to generate additional evidence or provide the final answer, based on the sufficiency of the current context.
    }
    \label{fig:overview}
\end{figure*}

\subsection{Empirical Study}
\label{sec:empirical_study}
To enable language models to generate relevant evidence existing in the external knowledge corpus, a natural approach is to apply FM-Index constraints over the entire corpus. However, our preliminary experiments reveal a critical limitation: while the initially generated evidence sequence usually appears relevant, later generated tokens often reveal that it has grounded to irrelevant documents under FM-Index constraints, resulting in incorrect evidence prediction. This phenomenon is known as \textbf{false pruning}, where relevant sequences are eliminated prematurely during beam search (see Appendix~\ref{app:false_pruning} for detailed analysis).

To quantify this issue, we conducted an empirical study. Figure~\ref{fig:empirical_study}(a) illustrates how the relevance calculated by bge-reranker-large between query and generated evidence prefix changes during the auto-regressive decoding process. The results show that compared to labeled evidence sequences, the prefix relevance under corpus FM-Index constraints experiences a significant decline, particularly severe within the first 13 tokens. When we restrict the FM-Index constraints to only relevant documents, this degradation is substantially reduced and evidence generation accuracy improves over different beam sizes (Figure~\ref{fig:empirical_study}(b)). This finding suggests that constraining the search space to a curated subset of relevant documents effectively mitigates false pruning, guiding the development of our strategies.

\section{RetroLLM: Retrieval in Generation}
In this section, we introduce RetroLLM, a unified LLM for RAG via auto-regressive decoding. The decoding process includes clue and evidence stages for retrieval and an answer generation stage. To achieve this, we describe the construction of constraints, clue generation, document scoring, and forward-looking constrained evidence generation.

\subsection{Hierarchical FM-Index Constraints}
Before model generation, we construct hierarchical FM-Indexes for different levels of constraints for clue and evidence generation stages, including: \textbf{(1) a corpus-level global FM-Index} $\mathcal{I}_c$ built from the entire corpus: $\mathcal{I}_c = \text{FM-Index}(\mathcal{C})$; and \textbf{(2) a document-level FM-Index manager} ${\mathcal{I}_d}$ built for each document: $\mathcal{I}_d = \text{FM-Index}(d): d \in \mathcal{C}$. 
The global index $\mathcal{I}_c$ is primarily used during the clue generation stage to ensure generated phrases exist in the corpus, while document-level indexes $\mathcal{I}_d$ are employed during evidence generation to constrain outputs to specific document $d$.

\subsection{Clue Generation and Document Scoring}
\label{sec:clue_doc_score}
As discussed in Section~\ref{sec:empirical_study}, generating evidence with relevant document FM-Indexes could reduce the decoding paths and enhance accuracy. Therefore, we propose that the LLM first predict key phrases, or ``clues,'' that are likely to appear in relevant documents to retrieve subsets of documents.

\textbf{Clue Generation.} Given a query $q$, we first generate a set of clues $\mathcal{C_\text{gen}}$ under corpus FM-Index constraints to predict key topics of relevant documents. For each clue $c_i \in \mathcal{C_\text{gen}}$, its generation probability can be formulated as:
\begin{equation}
P(c_i|q, c_{<i}, \mathcal{I}_c) = \prod\nolimits_{t=1}^{T_i} P(c_{i,t}|c_{i,<t}, q, c_{<i}, \mathcal{I}_c)
\end{equation}
where $c_{i,t}$ represents the $t$-th token of the $i$-th clue, $c_{i,<t}$ represents all previously generated tokens for the $i$-th clue, $\mathcal{I}_c$ is the corpus-level FM-Index, and $T_i$ is the length of the $i$-th clue. Clues are generated between the special tokens \red{<|clue|>} and \red{<|/clue|>}, separated by the special token \red{<|sep|>}.

With the predicted clues, we can obtain the appearance frequency $\text{CF}(c_i)$ of clue $c_i$ in the corpus based on the corpus FM-Index, along with $\text{DF}(c_i)$ which is the document frequency, and $\text{TF}(c_i, d)$ which is the term frequency in document $d$. Drawing inspiration from TF-IDF~\cite{bm25}, we assign higher weights to clues that appear less frequently in the corpus and are present in fewer documents. For a document $d$, we calculate the relevance score as:
\begin{equation}
\label{equ:gen_clue_score}
\mathcal{S}_\text{gen}(d) = \sum\nolimits_{i=1}^{|\mathcal{C_\text{gen}}|} w_i \cdot f(c_i, d),
\end{equation}
where $w_i$ is the weight of the $i$-th clue, defined as:
\begin{equation}
w_i = \log{\frac{N}{\text{CF}(c_i)}} + \log{\frac{N}{\text{DF}(c_i)}},
\end{equation}
and $f(c_i, d)$ scores the document $d$ for clue $c_i$:
\begin{equation}
f(c_i, d) = \log(1 + \text{TF}(c_i, d)),
\end{equation}
where $N$ is the total number of documents. Based on Equation~(\ref{equ:gen_clue_score}), we form the ranking list $R_1(d)$ by selecting the top-$k_{\text{gen}}$ documents from those containing at least one $c \in \mathcal{C}_{\text{gen}}$.

\textbf{Auxiliary Clues.} Although the generated clues could locate relevant documents intended by the model, they typically contain only 1-3 key phrases, which may limit comprehensive document recall. To enhance retrieval robustness, we obtain auxiliary clues by employing a sparse lexical model $f_\text{lex}$ that takes query $q$ as input and assigns importance weights to each word in its vocabulary $\mathcal{V}_\text{lex}$:
\begin{equation}
w_\text{lex}(v) = f_\text{lex}(q): v \in \mathcal{V}_\text{lex}.
\end{equation}
Subsequently, we select the top-$k_\text{aux}$ words from $\mathcal{V}_\text{lex}$ as auxiliary clues set $\mathcal{C_\text{aux}}$. Now we form a combined clue set $\mathcal{C_\text{all}} = \mathcal{C_\text{gen}} \cup \mathcal{C_\text{aux}}$. Additionally, we obtain a document ranking list $R_2(d)$ consisting of the top-$k_\text{lex}$ documents retrieved by $f_\text{lex}$.

\textbf{Rank Fusion.} The final candidate documents are determined by combining the ranking lists from both generated and expanded clues using weighted reciprocal rank fusion, which can be expressed as:
\begin{equation}
\mathcal{S}(d) = w_1 \sum_{r \in R_1(d)} \frac{1}{r} + w_2 \sum_{r \in R_2(d)} \frac{1}{r},
\end{equation}
where $w_1$ and $w_2$ are the respective weights for $R_1(d)$ and $R_2(d)$, $\frac{1}{r}$ represents the reciprocal rank score. Finally, the top-k ranked documents form the candidate set $\mathcal{D}_c$ for evidence generation.

\subsection{Forward-Looking Constrained Evidence Generation}
Now we have candidate documents, but a key challenge still remains: the model cannot foresee the relevance of future sequences when predicting the current token, making it difficult to decode tokens that lead to correct evidence sequences. To address this challenge, we propose a forward-looking constrained decoding strategy that enables the model to be aware of future sequence relevance.

Our strategy consists of three key components: (1) locating potential future windows that contain query-relevant information, (2) scoring the relevance of these windows, and (3) adjusting decoding logits based on future relevance. Give a candidate document set $\mathcal{D}_c$, $\mathcal{I}_{\mathcal{D}_c}$ is its FM-Indexes, the evidence generation process can be formulated as:
\begin{equation}
\label{equ:evidence_gen}
\begin{aligned}
& P(e_i|q, e_{<i}, \mathcal{C}_\text{all}, \mathcal{I}_{\mathcal{D}_c}) = \\ 
& \prod\nolimits_{t=1}^{T_i} P(e_{i,t}|e_{i,<t}, q, e_{<i}, \mathcal{C}_\text{all}, \mathcal{I}_{\mathcal{D}_c}, \mathcal{W}),
\end{aligned}
\end{equation}
where $e_i$ represents the $i$-th evidence sequence, $\mathcal{C}_\text{all}$ contains the generated clues, and $\mathcal{W}_\text{info}$ encapsulates future window information.

\textbf{Locate Future Windows.} 
First, we identify window sequences containing clues $\mathcal{C}_\text{all}$ in the candidate document set $\mathcal{D}_c$, as these contexts typically exhibit high relevance to the query. We obtain all future window sequences $\mathcal{W}_\text{raw}$ through document-specific FM-Indexes:
\begin{equation}
\mathcal{W}_\text{raw} = \bigcup_{d \in \mathcal{D}_c} \bigcup_{c \in \mathcal{C}_\text{all}} \text{Ext}(\mathcal{I}_d, \text{Loc}(\mathcal{I}_d, c), l_w).
\end{equation}
Here, $\text{Loc}(\mathcal{I}_d, c)$ locates clue positions in document $d$'s FM-Index $\mathcal{I}_d$, and $\text{Ext}(\mathcal{I}_d, p_c, l_w)$ extracts sequences of length $l_w$ around these positions. We then merge overlapping windows from $\mathcal{W}_\text{raw}$ to create the candidate future window set $\mathcal{W}$, with each merged window not exceeding length $l_\text{max}$.

\textbf{Future Window Relevance.} 
We employ a reranker model $f_\text{rel}$ to efficiently evaluate the relevance between each future window $w \in \mathcal{W}$ and the query $q$:
\begin{equation}
\mathcal{S}_\text{w}(w) = f_\text{rel}(q, w): w \in \mathcal{W}.
\end{equation}
Now for each $w \in \mathcal{W}$, we have its document source $d$, position $p_c$, and relevance score $\mathcal{S}_\text{w}(w)$.

\textbf{Logits Adjustment.} 
During decoding, we adjust token logits to favor sequences from highly relevant future windows. At each step, for allowed tokens $\mathcal{V}_\text{allowed}$ determined by FM-Index constraints, we locate each token's positions $\mathcal{P}_t$ and identify its corresponding future windows $\mathcal{W}_t$. The adjusted logits are computed as:
\begin{equation} 
\tilde{l}(t) = 
\begin{cases}
    l(t) + \lambda \cdot \max\limits_{w \in \mathcal{W}_t} \mathcal{S}_\text{w}(w), & \text{if } t \in \mathcal{V}_\text{allowed} \\
    -\infty, & \text{otherwise}
\end{cases}
\end{equation}
where $l(t)$ is the original logits and $\lambda$ controls the weight of future relevance. With logits adjustment, the token probability in Equation~(\ref{equ:evidence_gen}) is then computed as $P(e_{i,t}|...) = \text{softmax}(\tilde{l}(t))$.

The evidence generation process continues until sufficient information is collected, controlled by special tokens: \blue{<|evidence|>} begins evidence generation stage, \blue{<|sep|>} triggers the next evidence generation under constraints, while \blue{<|/evidence|>} signals completion and transitions to free generation.

\subsection{Answer Generation}
With the relevant evidences $\mathcal{E}$ generated, the model proceeds to generate the final answer to the original query $q$, which can be formulated as:
\begin{equation} P(a|q, \mathcal{C}_\text{gen}, \mathcal{E}) = \prod\nolimits_{t=1}^{T_a} P(a_t|a_{<t}, q, \mathcal{C}_\text{gen}, \mathcal{E}), 
\end{equation}
where $a$ is the generated answer sequence of length $T_a$, $a_t$ is the token at position $t$ in the answer, $a_{<t}$ denotes generated tokens before position $t$.

\begin{table*}[ht]
\centering
\caption{Overall performance on open-domain QA datasets, including single-hop and multi-hop QA tasks. The best results are in \textbf{bold} and the second are \underline{underlined}. Results from non-proprietary models are in \textcolor{gray!120}{gray} color.}
\label{tab:overall_qa_performance}
\setlength\tabcolsep{4.6pt}
\fontsize{8.9pt}{11pt}\selectfont
\begin{tabular}{p{2.25cm}ccccccccccccccc}
\toprule
 & \multicolumn{9}{c}{\textbf{In-domain Datasets}} & \multicolumn{6}{c}{\textbf{Out-of-domain Datasets}} \\
\cmidrule(lr){2-10} \cmidrule(lr){11-16}
\textbf{Method} & \multicolumn{3}{c}{\textbf{NQ}} & \multicolumn{3}{c}{\textbf{TriviaQA}} & \multicolumn{3}{c}{\textbf{HotpotQA}} & \multicolumn{3}{c}{\textbf{PopQA}} & \multicolumn{3}{c}{\textbf{2WIKI}} \\
\cmidrule(lr){2-4} \cmidrule(lr){5-7} \cmidrule(lr){8-10} \cmidrule(lr){11-13} \cmidrule(lr){14-16}
 & Acc & F1 & Tok & Acc & F1 & Tok & Acc & F1 & Tok & Acc & F1 & Tok & Acc & F1 & Tok \\
\midrule
\multicolumn{15}{l}{\textit{\textbf{Direct Generation}}} \\
Llama3-8B  & 27.6 & 30.1 & 50 & 56.1 & 60.2 & 52 & 21.2 & 26.5 & 56 & 24.2 & 26.4 & 43 & 20.9 & 24.3 & 54 \\
Mistral-7B & 30.4 & 25.2 & 57 & 58.8 & 58.6 & 57 & 27.0 & 23.6 & 65 & 25.8 & 25.2 & 45 & 36.5 & 18.7 & 58 \\
Qwen2.5-7B & 21.8 & 21.3 & 52 & 45.1 & 48.1 & 54 & 21.3 & 27.5 & 57 & 17.1 & 18.7 & 45 & 22.4 & 28.1 & 53 \\
ChatGPT & - & - & - & \textcolor{gray!120}{77.0} & \textcolor{gray!120}{52.9} & - & \textcolor{gray!120}{33.8} & \textcolor{gray!120}{24.0} & - & \textcolor{gray!120}{26.6} & \textcolor{gray!120}{13.2} & - & \textcolor{gray!120}{38.0} & \textcolor{gray!120}{21.3} & - \\
\midrule
\multicolumn{15}{l}{\textit{\textbf{Retrieval-augmented Generation}}} \\
Naive RAG       & \underline{52.4} & 41.1 & 919 & 69.3 & 65.9 & \underline{915} & \underline{37.8} & 35.8 & \underline{960} & 47.7 & 38.6 & 944 & \underline{38.7} & 21.7 & \underline{1000} \\
REPLUG          & 41.6 & 41.2 & \underline{903} & 65.4 & 66.5 & 939 & 27.8 & 31.7 & 965 & 38.2 & 37.0 & \underline{921} & 24.5 & 20.8 & 1007 \\
Self-RAG        & 41.8 & 45.2 & 1203 & 64.1 & 53.4 & 1267 & 32.1 & 29.6 & 1354 & 39.7 & 32.7 & 1236 & 30.3 & 25.7 & 1272 \\
IRCoT           & 49.6 & 45.9 & 1598 & 66.0 & 66.1 & 1715 & 37.3 & \underline{41.5} & 1842 & \underline{59.8} & \underline{45.6} & 1667 & 29.4 & \underline{32.4} & 1707 \\
Iter-RetGen     & 51.7 & \underline{48.4} & 3002 & \underline{71.0} & \underline{69.9} & 2461 & 37.2 & 39.0 & 2545 & 51.7 & \textbf{47.5} & 2509 & 29.2 & 21.5 & 2669 \\
Adaptive-RAG    & 50.5 & 46.6 & 946 & 65.1 & 65.6 & 958 & 37.1 & 39.1 & 2080 & 58.3 & 40.4 & 1681 & 32.1 & 28.4 & 2580 \\
\midrule
\multicolumn{15}{l}{\textit{\textbf{Retrieval within Generation}}} \\
\rowcolor[RGB]{236,244,252} 
RetroLLM (Ours)        & \textbf{61.6} & \textbf{49.8} & \textbf{302} & \textbf{74.3} & \textbf{72.8} & \textbf{287} & \textbf{61.9} & \textbf{47.2} & \textbf{607} & \textbf{65.7} & 43.0 & \textbf{355} & \textbf{48.9} & \textbf{36.2} & \textbf{661} \\
\bottomrule
\end{tabular}
\end{table*}

\subsection{Training of RetroLLM}
\label{sec:training}
Since RetroLLM's entire RAG process is one-pass and auto-regressive, we can construct target sequences for supervised fine-tuning to achieve joint learning of retrieval and generation tasks.

\textbf{Training Data Construction.}
We simulate the model's inference process to construct training data. For each QA pair $(q, a)$, we: (1) Use a sparse retriever to obtain clues $\mathcal{C}_\text{aux}$ and retrieve relevant documents. (2) Locate sentences containing $c \in \mathcal{C}_\text{aux}$ within the documents. (3) Apply a reranker to select the top-$k_e$ relevant evidences. (4) Identify evidences that both contain the answer $a$ and are confirmed by an LLM to genuinely answer the query $q$. (5) Select the top-$k$ evidences up to the first relevant one. (6) For target clues, we utilize an LLM to extract key entities from the query and relevant evidences. 
An example of the output format is illustrated in Figure~\ref{fig:overview}, and additional details are provided in Appendix~\ref{app:training}.

\textbf{Model Optimization.}
Since evidence is typically longer compared to clues and answer, we mask out 80\% of the tokens in the middle of each target evidence. We employ the standard next token prediction loss as follows:
\begin{equation}
\begin{aligned}
\mathcal{L} &= -\sum\nolimits_{t=1}^{T_c + T_e} \log P(x_t|x_{<t}, q; \theta)\\ &- \gamma \sum\nolimits_{t=1}^{T_a} \log P(y_t|y_{<t}, x, q; \theta),
\end{aligned}
\end{equation}
where $\theta$ represents the parameters of RetroLLM, $x$ and $y$ represent the target sequence of clues + evidences and answer respectively, and $\gamma$ is the weight for the answer loss.

\section{Experimental Settings}
\subsection{Datasets and Evaluation Metrics}
We conduct experiments on five open-domain QA datasets, including single-hop QA: NQ~\cite{nq}, TriviaQA~\cite{triviaqa}, PopQA~\cite{popqa}, and multi-hop QA: HotpotQA~\cite{hotpotqa}, 2WikiMultiHopQA (2WIKI)~\cite{2wiki}. See Appendix~\ref{app:datasets} for detailed statistics. For evaluation metrics, we use Accuracy (Acc), F1 score, and token count (Tok) to assess the quality of generated answers as well as the total number of input and output tokens consumed by LLMs.

\subsection{Baselines}
The baseline methods include two types: (1) Direct generation: This includes open-source models Llama3-8B~\cite{llama3}, Mistral-7B~\cite{mistral_7b}, and Qwen2.5-7B~\cite{qwen2}, and the non-proprietary model ChatGPT~\cite{chatgpt} with results taken from~\cite{2408_raglab}. (2) Retrieval-augmented generation: This includes Naive RAG method and several complex RAG methods, including REPLUG~\cite{replug}, Self-RAG~\cite{self-rag}, IRCoT~\cite{ircot}, Iter-RetGen~\cite{iterretgen}, and Adaptive-RAG~\cite{adaptive-rag}. For a fair comparison, all RAG baselines use the E5-base-en~\cite{e5} retriever, and all LLMs are instruction-tuned with 7B or 8B parameters.

\subsection{Implementation Details}
Our knowledge source is based on the Wikipedia dump from December 20, 2018, in alignment with DPR~\cite{dpr}. We use Mistral-7B-Instruct as Backbone LLM. We limit the maximum number of evidence for single-hop and multi-hop QA to 5 and 10, respectively. We set $w_1$ and $w_2$ to 1 and 2, respectively, and  $\lambda$ to 100. For efficient model training, we employ LoRA~\cite{lora}, setting training epochs to 3 and $\gamma$ to 2. We use SPLADE-v3~\cite{splade-v3} for clue expansion and use BGE-reranker-base~\cite{bge} as the scoring model for window sequences. We implement FM-Index based on the sdsl-lite library~\cite{sdsl_lite}. Refer to Appendix~\ref{app:implementation} for more details.

\subsection{Experimental Results}

\paragraph{Overall Performance}
We evaluate RetroLLM's overall downstream performance using NQ, TriviaQA, and HotpotQA for in-domain tasks, and PopQA and 2WIKI for out-of-domain tasks. The results are presented in Table~\ref{tab:overall_qa_performance}. We could found that: 
{(1) RAG methods generally outperform direct generation methods (except for non-proprietary ChatGPT)}, highlighting the knowledge-intensive nature of these tasks that need retrieval augmentation.
\textbf{(2) RetroLLM outperforms RAG methods across both in-domain and out-of-domain tasks.} This highlights the effectiveness of our unified RAG framework in mastering both evidence retrieval and answer generation, while also demonstrating strong generalization capabilities to unfamiliar domains, which is a crucial and challenging aspect for existing generative retrieval methods.
\textbf{(3) RetroLLM significantly reduces token consumption (``Tok'') across all datasets compared to RAG methods.} On average, we use approximate 2.1x fewer tokens than Naive RAG and 6x fewer than Iter-RetGen. This efficiency is attributed to RetroLLM's capability to retrieve fine-grained evidence and dynamically decide the amount of retrieved evidence.

\paragraph{Analysis of Retrieval Performance}
\label{par:retrieval_analysis}
We also analyze the retrieval performance of RetroLLM compared to sparse and dense retrieval baselines, as shown in Table~\ref{tab:retrieval_performance}. 
(1) For single-hop QA tasks, RetroLLM demonstrates superior accuracy on R@1, thanks to the design of clues and future windows, which help precisely locate the relevant evidence. However, the R@5 is lower than strong baselines like E5, as it employs flexible retrieval and uses fewer passages on average (3.29 vs. 5 for baselines). 
(2) For multi-hop QA tasks, RetroLLM shows superior accuracy compared to all other methods for both R@1 and R@5, while utilizing a smaller average number of 4.24 retrieved passages. 
(3) Notably, the naive generative retrieval method using constrained beam search performs poorly on all metrics, further validating the severity of false pruning, as discussed in Section~\ref{sec:empirical_study}.

\begin{table}[!t]
\centering
\caption{Analysis of retrieval performance of RetroLLM, compared with sparse, dense, and generative retrieval methods. We report average performance on three single-hop and two multi-hop QA datasets.}
\label{tab:retrieval_performance}
\setlength\tabcolsep{4pt}
\fontsize{8.6pt}{10.4pt}\selectfont
\begin{tabular}{lcccccc}
\toprule
\multirow{2}[2]{*}{\textbf{Method}} & \multicolumn{3}{c}{\textbf{Single-hop QA}} & \multicolumn{3}{c}{\textbf{Multi-hop QA}} \\
\cmidrule(lr){2-4} \cmidrule(lr){5-7}
 & R@1 & R@5 & Num & R@1 & R@5 & Num \\
\midrule
BM25 & 37.8 & 56.3 & 5 & 26.9 & 43.1 & 5 \\
SPLADE-v3 & 50.6 & 69.7 & 5 & 27.5 & 42.9 & 5 \\
E5 & 54.3 & \textbf{74.3} & 5 & 26.9 & 45.9 & 5 \\
BGE & 53.3 & 72.8 & 5 & 27.4 & 46.8 & 5 \\
Naive Constrain & 15.7 & 31.7 & 5 & 10.6 & 20.3 & 5 \\
\rowcolor[RGB]{236,244,252} 
RetroLLM & \textbf{56.6} & 67.9 & \textbf{3.29} & \textbf{29.3} & \textbf{49.6} & \textbf{4.24} \\
\bottomrule
\end{tabular}
\end{table}

\paragraph{Ablation Study}
\label{sec:ablation}
Table~\ref{tab:ablation_study} presents the results of the ablation study, evaluating the effectiveness of each component of RetroLLM. It can be observed that:
(1) Removing the future window, clue generation, and clue expansion all lead to performance degradation, demonstrating the effectiveness of these specially designed components, as they play an important role in alleviating the false pruning problem in prefix constraint-based method.
(2) Adopting the most basic constrained evidence generation method results in the lowest performance, even lower than without constraints, demonstrating the severity of false pruning.
(3) Without constraints, while the model avoids the false pruning problem, its performance still notably decreases due to the inability to utilize external knowledge.

\begin{figure}[!t]
\begin{subfigure}{.495\linewidth}
    \centering
    \includegraphics[width=\linewidth]{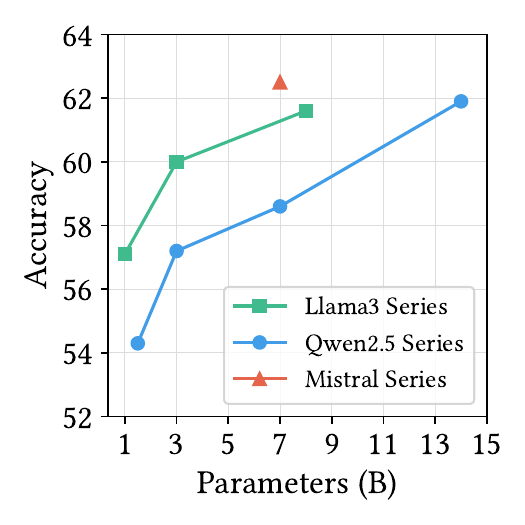}
    \caption{Parameters vs. Accuracy}
\end{subfigure}
\begin{subfigure}{.495\linewidth}
    \centering
    \includegraphics[width=\linewidth]{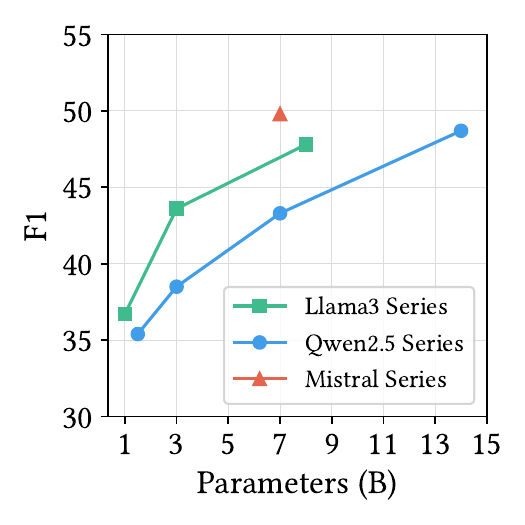}
    \caption{Parameters vs. F1}
\end{subfigure}
\caption{Impact of performance with different base LLMs, reporting average performance on five datasets.}
\label{fig:impact_base_llm}
\end{figure}

\begin{table}[!t]
\centering
\caption{Ablation Studies of RetroLLM, considering in-domain and out-of-domain performance.}
\label{tab:ablation_study}
\setlength\tabcolsep{5.9pt} 
\fontsize{8.6pt}{10.4pt}\selectfont
\begin{tabular}{p{2.7cm}p{0.65cm}<{\centering}p{0.65cm}<{\centering}p{0.7cm}<{\centering}p{0.7cm}<{\centering}}
\toprule
\multirow{2}[2]{*}{\textbf{Method}} & \multicolumn{2}{c}{\textbf{In-domain}} & \multicolumn{2}{c}{\textbf{Out-of-domain}} \\
\cmidrule(lr){2-3} \cmidrule(lr){4-5}
 & Acc & F1 & Acc & F1 \\
\midrule
\rowcolor[RGB]{236,244,252} 
RetroLLM & \textbf{66.0} & \textbf{56.6} & \textbf{57.3} & \textbf{39.6} \\
\midrule
w/o Future Window & 44.3 & 43.2 & 40.9 & 33.8 \\
w/o Clue Generation & 60.6 & 52.1 & 56.4 & 38.1 \\
w/o Clue Expansion & 49.6 & 45.1 & 44.1 & 35.4 \\
w/ Naive Constraints & 27.2 & 28.0 & 21.8 & 20.7 \\
w/o Constraints & 41.6 & 43.0 & 31.6 & 28.1 \\
\bottomrule
\end{tabular}
\end{table}

\paragraph{Impact of Different Base LLMs}
\label{par:base_model_analysis}
To evaluate the performance of RetroLLM using different backbone LLMs with varying parameter sizes, we conducted experiments using the Mistral, Llama3, and Qwen2.5 series, with parameters ranging from 1B to 14B. The results are shown in Figure~\ref{fig:impact_base_llm}. We observe that: (1) As the parameter size increases, RetroLLM's performance steadily improves, aligning with the scaling law; (2) There are slight performance differences across the different models (Mistral, Llama3, Qwen2.5), with Mistral generally outperforming Llama3, which in turn outperforms Qwen2.5. Nonetheless, all models confirm the effectiveness of RetroLLM (see Appendix~\ref{app
} for detailed results).

\begin{figure}[!t]
\begin{subfigure}{.495\linewidth}
    \centering
    \includegraphics[width=\linewidth]{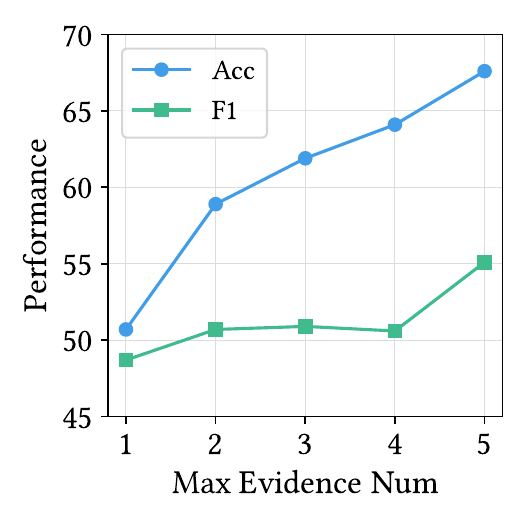}
    \caption{Single-hop QA}
\end{subfigure}
\begin{subfigure}{.495\linewidth}
    \centering
    \includegraphics[width=\linewidth]{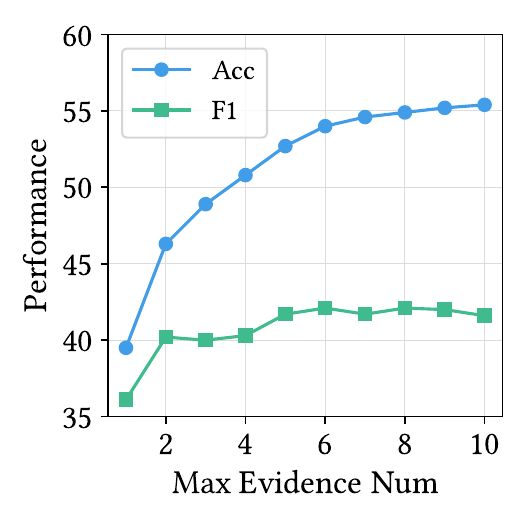}
    \caption{Multi-hop QA}
\end{subfigure}
\caption{Impact on maximum number of generated evidence, reporting average performance on single-hop and multi-hop QA tasks.}
\label{fig:impact_evidence_num}
\end{figure}

\paragraph{Impact of Generated Evidence Quantity}
\label{par:evidence_num_analysis}
Since RetroLLM can dynamically determine the number of evidence to retrieve, we investigated the effect of different maximum retrieval quantities on performance, with results shown in Figure~\ref{fig:impact_evidence_num}. When retrieving up to 1-5 evidence, performance continues to improve as the number of retrieved pieces increases, suggesting that more evidence contributes to stronger performance in these tasks. However, for multi-hop QA, performance stabilizes around 6 evidence, as more evidence can bring in both useful and distracting information, thereby limiting further performance gains.

\paragraph{Analysis of Efficiency}
We also evaluated the efficiency of RetroLLM, considering query latency, token count, and overall performance (see Table~\ref{tab:efficiency_analysis}). We found that: \textbf{(1) Latency:} RetroLLM is slightly slower than Naive RAG but significantly faster than other more complex RAG methods. \textbf{(2) Token Count:} RetroLLM requires fewer input tokens as it processes only the query, unlike baselines that include retrieved passages. While output tokens are slightly higher due to fine-grained generated evidence. Total token count is significantly reduced due to more precise retrieval granularity. \textbf{(3) Performance:} RetroLLM achieves better results, providing an improved cost-efficiency balance.

\section{Related Work}
\paragraph{Retrieval-augmented Generation}
Retrieval-augmented generation (RAG) improves generation quality by incorporating relevant context from external knowledge bases, which typically employ a separate dense retriever~\cite{ragsurvey, slimplm_jiejun, bider_jiajie, 2411_htmlrag, trustworthy_survey}. Based on training approaches, current RAG systems fall into three categories: (1) Directly prompt of generative models with retrieved context~\cite{self-ask, ircot}; (2) Separately training of retriever and/or generator~\cite{dpr, self-rag, 2405_spring, 2406_dpa_rag, 2410_vif_rag}; and (3) Jointly training of retriever and generator~\cite{rag, emdr2}. However, joint training faces challenges due to the architectural differences between retrieval and generation, as well as the need for updating document indices during training. Some approaches aim to unify dense retrieval and generation within a single model, including GritLM~\cite{2402_gritlm} and OneGen~\cite{2409_onegen}. However, GritLM operates as two distinct models with separate attention mechanisms that share parameters, while OneGen still relies on retrieving passage chunks as input for subsequent generation.

\begin{table}[!t]
\centering
\caption{Efficiency Analysis of RetroLLM, comparing query latency, number of tokens and performance (\# P).}
\label{tab:efficiency_analysis}
\setlength\tabcolsep{3.6pt}
\fontsize{8.6pt}{10.5}\selectfont
\begin{tabular}{p{1.6cm}ccccccc}
\toprule
\multirow{2}[2]{*}{\textbf{Method}} & \multicolumn{3}{c}{\textbf{Latency (ms)}} & \multicolumn{3}{c}{\textbf{Token Num}} & \multirow{1}[1]{*}{\textbf{\# P}} \\
\cmidrule(lr){2-4} \cmidrule(lr){5-7}\cmidrule(lr){8-8}
 & Retr & Gen & Total & In & Out & Total & F1 \\
\midrule
Naive RAG & \textbf{54} & \textbf{528} & \textbf{582} & 902 & \textbf{17} & 919 & 41.1 \\
SelfRAG & 89 & 3180 & 3269 & 1096 & 107 & 1203 & 45.2 \\
Iter-RetGen & 274 & 2058 & 2332 & 2963 & 39 & 3002 & 48.4 \\
IRCoT & 83 & 1759 & 1842 & 1535 & 63 & 1598 & 46.6 \\
\rowcolor[RGB]{236,244,252} 
RetroLLM & - & - & 786 & \textbf{18} & 297 & \textbf{315} & \textbf{49.8} \\
\bottomrule
\end{tabular}
\end{table}

\paragraph{Generative Retrieval}
Generative retrieval (GR) retrieves by directly generating document identifiers (DocIDs) without the need for traditional document indices~\cite{metzler2021rethinking}. Research in this area focuses on: (1) DocID design, including numeric-based DocIDs~\cite{dsi, nci, se-dsi, lmindexer, ripor} and text-based DocIDs~\cite{genre, seal, ultron, corpusbrain, autotsg, minder, genrrl}; (2) DocID memorization strategies, including pseudo-query data augmentation~\cite{dsiqg}, incorporating ranking feedback~\cite{ltrgr, listgr}, and learnable DocIDs~\cite{genret, novo, asi}. However, these methods mainly focus on optimizing retrieval tasks, without considering its connections with downstream tasks. Even though UniGen~\cite{unigen} and CorpusLM~\cite{2402_corpuslm} address downstream tasks, they still require mapping the generated DocIDs to the corresponding documents before feeding them into the generator. While RICHES~\cite{2407_riches} attempts to streamline this process but fails to solve the false pruning issue, which leads to suboptimal downstream task performance.

\section{Conclusion}
In this paper, we introduced RetroLLM, a framework that unifies retrieval and generation into a single process, allowing language models to directly generate evidence from a corpus with FM-Index constraints. This approach eliminates the need for separate retrievers and reduces redundant input. To improve evidence accuracy, we proposed hierarchical FM-Index constraints and a forward-looking decoding strategy, helping the model focus on relevant information. Experiments show that RetroLLM outperforms existing methods on open-domain QA tasks, marking a step towards a new era of generative retrieval-augmented generation.


\section{Limitations}
While RetroLLM demonstrates strong performance across various open-domain QA scenarios, it has several limitations that present opportunities for future research:

(1) To improve the robustness of the model generated clues, we still need to perform clue expansion to ensure the system's superior performance, as discussed in Section~\ref{sec:ablation}. This prevents a fully end-to-end optimization of the RAG task. Future work could focus on designing mechanisms that eliminate this need, enabling complete end-to-end RAG optimization.

(2) In terms of efficiency, RetroLLM outperforms most complex RAG methods in query latency. However, it is slightly slower than Naive RAG, as the generated evidence results in more output tokens despite being fine-grained and short. Drawing on the concept of speculative decoding~\cite{23_icml_Speculative_Decoding, 23_emnlp_Speculative_Decoding}, future improvements could involve using a smaller language model during the constrained evidence generation phase and switching to a larger model during answer generation. This approach could enhance RetroLLM's efficiency, comprehensively surpassing existing RAG methods in both performance, latency, and flexibility.

(3) RetroLLM currently only considers the unification of evidence retrieval and final answer generation. It would be worth exploring the incorporation of more model reasoning processes within RetroLLM's single generation step, such as query intent analysis, question decomposition, retrieval necessity assessment, evidence relevance judgment, and answer generation with source citations. This integration would contribute to building a more comprehensive unified RAG system with just a single LLM.

\bibliography{main}

\clearpage
\appendix

\onecolumn
\section*{Appendix}
\startcontents[sections]
\printcontents[sections]{l}{1}{\setcounter{tocdepth}{3}}
\twocolumn

\clearpage

\section{The FM-Index}
\label{app:fm_index}
The FM-Index~\cite{fm_index}, which stands for Full-text index in Minute space, is a space-efficient data structure designed for indexing large text corpora, combining the Burrows-Wheeler Transform (BWT) and run-length encoding. It enables fast substring searching while providing substantial compression, making it particularly useful in applications such as prefix-constrained decoding.

\subsection{Data Structure}

The FM-Index is based on the Burrows-Wheeler Transform (BWT)~\cite{bwt_transform}. The BWT of a string $S$ is computed by sorting all cyclic rotations of $S$ lexicographically and then taking the last column of the sorted rotations. This transformation rearranges the characters of the string in a way that enhances its compressibility, which is key to the FM-Index's space efficiency.

Formally, for a string $S = S_1 S_2 \dots S_n$, the BWT, denoted as $\text{BWT}(S)$, is obtained by sorting the cyclic rotations of $S$ lexicographically and taking the last character of each rotation. Let $\mathcal{R}(S)$ denote the set of all cyclic rotations of $S$, sorted in lexicographical order:
\begin{equation}
\mathcal{R}(S) = \left\{ \sigma_1, \sigma_2, \dots, \sigma_n \right\}
\end{equation}
where $\sigma_i$ denotes the $i$-th rotation of $S$. The BWT of $S$ is then the string formed by the last characters of these sorted rotations:
\begin{equation}
\text{BWT}(S) = \left( \sigma_1[n], \sigma_2[n], \dots, \sigma_n[n] \right)
\end{equation}

The FM-Index stores only two columns from the BWT matrix: the first (F) and last (L) columns. These columns capture the relative order of the characters in all cyclic rotations of $S$. More precisely:
- The first column (F) contains the sorted characters of the text $S$.
- The last column (L) contains the last character of each of the cyclic rotations of $S$.

Additionally, to enable efficient searching, the FM-Index uses additional data structures, such as the Wavelet Tree, to store the L column efficiently. This allows for fast rank and select operations, which are essential for searching.

\subsection{Supporting Functions}

\paragraph{Backward Search}
The core feature of the FM-Index is the backward search, which allows for efficient substring searching. Given a substring $P = p_1 p_2 \dots p_k$, the backward search locates all occurrences of $P$ in the original string by iteratively searching through the first (F) and last (L) columns.

The backward search proceeds by iterating through each character of $P$ from right to left. Initially, the search interval spans the entire text $S$, represented as the BWT matrix. In each step, we examine the current character $p_i$ of the pattern and update the search interval by examining the corresponding entries in the F and L columns. Specifically:
1. We find all occurrences of $p_i$ in the last column (L).
2. We update the search range in the first column (F) to include only the rows corresponding to the occurrences of $p_i$.

This process is repeated for each character of the pattern, refining the search interval until all occurrences of the substring are found.

The time complexity of the backward search is $O(k \log |V|)$, where $k$ is the length of the pattern $P$ and $|V|$ is the size of the alphabet. This is because each iteration of the search takes $O(\log |V|)$ time, and there are $k$ iterations corresponding to the length of the pattern.

\paragraph{Count}
The occurrence count function, denoted as $\text{occ}(P)$, counts how many times a pattern $P$ occurs in the original text. The occurrence count is closely related to the backward search. After performing a backward search for a pattern $P$, the occurrence count is simply the size of the resulting search interval. This interval represents all occurrences of $P$ in the text.

Since computing the occurrence count requires performing a backward search, the time complexity of this operation is also $O(k \log |V|)$, where $k$ is the length of the pattern and $|V|$ is the size of the alphabet.

\paragraph{Locate}
The locate function, denoted as $\text{locate}(P)$, returns the positions in the original text where the pattern $P$ occurs. This function works by using the results of the backward search to determine the positions of the occurrences. Specifically, once the search interval is determined through backward search, the locate function maps the rows of the interval back to positions in the original text. This mapping is achieved by using the F column, which contains the sorted characters of the text.

The time complexity of the locate function is $O(k \log |V|)$, as it involves performing a backward search for the pattern $P$.

\paragraph{Extract}
Given the start and end positions of a substring in the BWT matrix, the extract text function reconstructs the corresponding text substring. The process works by tracing the positions of the characters in the substring through the F and L columns in reverse order. Starting from the end position, the algorithm follows the reverse of the backward search to find the characters in the substring and reconstruct the text.

This operation runs in $O(k)$, where $k$ is the length of the substring to be extracted. The reason for this is that we need to perform $k$ steps to extract a substring of length $k$, with each step involving simple lookups in the F and L columns.

\subsection{Examples}

To better understand how the FM-Index works, let us examine a concrete example using the string $S = \text{"banana\$"}$, where \$ serves as the end-of-string marker. The first step in constructing the FM-Index is to generate the Burrows-Wheeler Transform. This is accomplished by creating all possible cyclic rotations of the input string and sorting them lexicographically. For our example string, the sorted rotations form a matrix where each row represents one rotation:
$$
\begin{array}{c|c|c}
\hline
\text{Sorted Rotations} & \text{F Col.} & \text{L Col.} \\
\hline
\text{\$banana} & \text{\$} & \text{a} \\
\text{a\$banan} & \text{a} & \text{n} \\
\text{ana\$ban} & \text{a} & \text{n} \\
\text{anana\$b} & \text{a} & \text{b} \\
\text{banana\$} & \text{b} & \text{\$} \\
\text{na\$bana} & \text{n} & \text{a} \\
\text{nana\$ba} & \text{n} & \text{a} \\
\hline
\end{array}
$$
The Burrows-Wheeler Transform of $S$ is then obtained as the last column L: $\text{BWT}(S) = \text{"annb\$aa"}$. The FM-Index maintains this last column L along with the first column F = "\$aaabnn", which contains the lexicographically sorted characters of $S$. These two columns, combined with auxiliary data structures for efficient rank and select operations, form the core of the FM-Index.

Given a pattern such as $P = \text{"ana"}$, we can perform a backward search from right to left. Starting with the last character 'a', we determine its occurrence range in the F column using the cumulative count \( C[a] = 1 \) (since only '\$' precedes 'a'). The rank of 'a' up to position 7 in L is 3, updating the search interval to [2, 4]. Next, processing the character 'n', with \( C[n] = 5 \), we find that the rank of 'n' up to position 4 in L is 2, refining the search interval to [6, 7]. Finally, processing the first character 'a', we use \( C[a] = 1 \) and find that the rank of 'a' up to position 7 in L is 3, resulting in a final search interval of [3, 4].

Upon identifying the final search interval [3, 4] in the L column, we examine the corresponding characters, which are 'n' and 'b'. These characters represent the ones that precede the pattern "ana" in the original string \( S \). Mapping these to the positions in the F column reveals that the possible characters following "ana" in \( S \) are 'n' and '\$'. Specifically, in "banana\$", the substring "ana" is followed by 'n' in the first occurrence and by '\$' in the second occurrence. Therefore, the backward search correctly identifies 'n' and '\$' as the allowable next characters after the prefix "ana", validating the FM-Index derivation process.

\section{False Pruning in Constrained Decoding}
\label{app:false_pruning}
In Section~\ref{sec:empirical_study}, we conducted empirical studies revealing that false pruning is a significant issue in constrained decoding. This section delves deeper into understanding this problem.

\subsection{What is False Pruning?}
False pruning occurs when the search process incorrectly eliminates branches that could contain the optimal solution, preventing the algorithm from identifying the true best outcome. Specifically, in prefix-constrained decoding for language models, false pruning involves incorrectly discarding candidate tokens that meet the prefix constraint but might contribute to the optimal solution~\cite{autotsg}.

Consider the question: ``\texttt{What is the capital of France?}'' The correct evidence in the corpus is ``\texttt{The capital city of France is Paris.}'' During decoding with the prefix constraint ``\texttt{The capital}'', if the model selects ``\texttt{of}'' instead of ``\texttt{city}'', a critical issue arises. Suppose the corpus lacks direct statements like ``\texttt{The capital of France}'' but includes irrelevant examples such as ``\texttt{The capital of America}''. In this scenario, even though the path starting with ``\texttt{The capital of}'' could potentially lead to the correct answer about Paris, the model is constrained by the corpus to decode only irrelevant evidence like ``\texttt{The capital of America is Washington D.C.}''.

This situation exemplifies false pruning: the correct solution path is erroneously removed during beam search or sampling, despite being valid under the prefix constraint. This happens because intermediate token choices steer the model toward contexts where it cannot effectively retrieve the target information about France's capital. Such failures illustrate how the local, token-by-token nature of decoding can clash with prefix constraints, causing the model to follow suboptimal paths and ultimately fail to generate the correct evidence.

\subsection{What Causes False Pruning?}
For auto-regressive decoding models, false pruning arises primarily due to two factors:

\textbf{Excessive Prefix Choices:} In large corpora, candidate sequences present a vast number of prefix options initially. The model can generate nearly any short prefix it wants, making it challenging to predict the correct one.
    
\textbf{Limited Future Awareness:} Even with fewer prefix choices, the model cannot anticipate future content beyond the current token decision. This limitation makes it difficult to select tokens that lead to the correct evidence.

\subsection{How to Mitigate False Pruning?}
Addressing the root causes of false pruning involves implementing strategies that either narrow the prefix choices or enhance the model's foresight during decoding.

\textbf{Reducing Prefix Choices:} One effective method is to limit the number of prefix options. Our approach employs clue generation to identify a relevant subset of documents, followed by decoding evidence within this constrained set. This reduction significantly decreases the prefix choices, mitigating the risk of false pruning.

\textbf{Enhancing Future Relevance Awareness:} Another strategy is to provide the model with information about the relevance of future sequences. In our method, we identify the clue's position within the document and utilize the surrounding text as future windows. By guiding the language model to generate relevant evidence based on these windows and their relevance scores, we improve the model's ability to connect to the target information.

\textbf{Set-Based Decoding:} Some generative retrieval methods adopt set-based decoding strategies~\cite{autotsg,PAG}, which bypass the issues inherent in auto-regressive decoding by directly generating sets of terms. These methods are suitable for retrieval tasks that involve decoding document identifiers to fetch corresponding documents. However, they are not applicable to our evidence generation task, where the goal is to generate meaningful evidence directly rather than retrieve identifiers.




\begin{table}[!t]
\centering
\caption{Detailed statistics of datasets and retrieval corpus utilized in our experiments.}
\label{tab:dataset_info}
\setlength\tabcolsep{8pt}
\fontsize{8.6pt}{10.4}\selectfont
\begin{tabular}{llrr}
\toprule
\textbf{Task} & \textbf{Dataset} & \textbf{\# Train} & \textbf{\# Test} \\
\midrule
Single-hop QA & NQ & 79,168 & 3,610 \\
Single-hop QA & TriviaQA & 78,785 & 11,313 \\
Single-hop QA & PopQA & / & 14,267 \\
Multi-hop QA & HotpotQA & 90,447 & 7,405 \\
Multi-hop QA & 2WIKI & / & 12,576 \\
\midrule
\textbf{Retrieval Corpus} & \textbf{\# Passages} & \multicolumn{2}{r}{\textbf{\# Documents}} \\
\midrule
Wikipedia & 21,015,324 & \multicolumn{2}{r}{3,232,907} \\
\bottomrule
\end{tabular}
\end{table}

\section{Datasets}
\label{app:datasets}

\subsection{Details of Datasets}
In our experiments, we utilize a variety of question answering (QA) datasets to evaluate both single-hop and multi-hop reasoning capabilities. For single-hop QA, we employ the Natural Questions (NQ) \cite{nq} dataset, TriviaQA \cite{triviaqa}, and PopQA \cite{popqa}, which provide a diverse range of factual questions requiring straightforward retrieval and answer extraction. For multi-hop QA, we use HotpotQA \cite{hotpotqa}, which necessitates reasoning across multiple documents, and 2WIKI \cite{2wiki}, a dataset designed to test more complex multi-hop reasoning scenarios. These datasets are selected to cover a broad spectrum of QA challenges, ensuring a comprehensive evaluation of model's retrieval and reasoning capability.

\subsection{Statistics}
Table~\ref{tab:dataset_info} presents detailed statistics of the datasets and the retrieval corpus used in our study. For single-hop QA tasks, NQ consists of 79,168 training samples and 3,610 test samples, while TriviaQA has 78,785 training samples and 11,313 test samples. PopQA is used solely for testing, with 14,267 samples. In the multi-hop QA category, HotpotQA includes 90,447 training samples and 7,405 test samples, and 2WIKI provides 12,576 test samples without a training set. The retrieval corpus comprises the Wikipedia dataset, containing 21,015,324 passages and 3,232,907 documents. These statistics highlight the extensive scale of our experimental setup, facilitating robust training and evaluation of our models.



\section{Implementation Details}
\label{app:implementation}

\subsection{Implementation Details for Baselines}
All RAG baselines are implemented based on the FlashRAG framework, which is an open-source retrieval-augmented generation toolkit~\cite{flashrag_jiajie}. For Self-RAG~\cite{self-rag}, we use the trained \texttt{selfrag-llama2-7B} checkpoint. For all other baselines, we use \texttt{Mistral-7B-Instruct} as the backbone model, aligning with our RetroLLM. All hyper-parameter configurations are set to default in FlashRAG.

\subsection{Implementation Details for Naive Constrained Generation}
For the naive approach to constrained beam evidence generation, we set \texttt{num\_beams} = 5, \texttt{num\_beam\_groups} = 5, and \texttt{diversity\_penalty} = 1.0 for constrained beam search. The \texttt{num\_beam\_groups} and \texttt{diversity\_penalty} parameters are crucial; without setting these two parameters, the sequences generated by each beam would be highly similar, leading to a significant decrease in evidence accuracy. These parameters ensure diversity among the multiple generated sequences and sort the \texttt{beam\_size} generated evidences from high to low according to the generation probability of the language model, so that more relevant evidence can be ranked ahead.

For cases where an answer needs to be generated, we continue to freely generate the answer without constraint after each beam, and the final answer given is the answer generated after the top-1 sequence. The specific input and output formats are shown in~\ref{app:case_study}.

\subsection{Implementation Details for RetroLLM}
The implementation of RetroLLM mainly includes FM-Index building, training, and inference. All experiments are conducted on 8 NVIDIA A800-80GB GPUs and an Intel(R) Xeon(R) Platinum 8358 CPU @ 2.60GHz with 64 cores.

\subsubsection{FM-Index Building}
We implement the FM-Index data structure based on the SDSL-lite (Succinct Data Structure Library) framework~\cite{sdsl_lite}, which is an efficient C++ template library specifically designed for implementing compressed data structures. We then implemented the functionalities used in this paper at the C++ level, including prefix locating, finding allowed next tokens, counting occurrences, extracting sequences, etc. We also built and stored an FM-Index Manager on the C++ side to map given DocIDs to their corresponding document FM-Indexes. To allow Python code to call these C++ implementations, we used the SWIG (Simplified Wrapper and Interface Generator) tool~\cite{beazley1996swig}.

\begin{table*}[ht]
\centering
\caption{Detailed retrieval performance on five open-domain QA datasets, comparing sparse, dense, and generative approaches. The best results are highlighted in \textbf{Bold}.}
\label{tab:retrieval_performance_app}
\setlength\tabcolsep{3.6pt}
\fontsize{8.9pt}{10.5pt}\selectfont
\begin{tabular}{p{2.2cm}ccccccccccccccc}
\toprule
 & \multicolumn{9}{c}{\textbf{In-domain Datasets}} & \multicolumn{6}{c}{\textbf{Out-of-domain Datasets}} \\
\cmidrule(lr){2-10} \cmidrule(lr){11-16}
\textbf{Method} & \multicolumn{3}{c}{\textbf{NQ}} & \multicolumn{3}{c}{\textbf{TriviaQA}} & \multicolumn{3}{c}{\textbf{HotpotQA}} & \multicolumn{3}{c}{\textbf{PopQA}} & \multicolumn{3}{c}{\textbf{2WIKI}} \\
\cmidrule(lr){2-4} \cmidrule(lr){5-7} \cmidrule(lr){8-10} \cmidrule(lr){11-13} \cmidrule(lr){14-16}
 & R@1 & R@5 & Num & R@1 & R@5 & Num & R@1 & R@5 & Num & R@1 & R@5 & Num & R@1 & R@5 & Num \\
\midrule
\multicolumn{15}{l}{\textit{\textbf{Sparse Retrieval}}} \\
BM25 & 24.1 & 46.2 & 5 & 49.6 & 68.5 & 5 & 31.2 & 48.7 & 5 & 39.6 & 54.3 & 5 & 22.6 & 37.5 & 5 \\
SPLADE-v3 & 45.4 & 68.0 & 5 & 58.8 & 75.9 & 5 & 32.9 & 45.3 & 5 & 47.6 & 65.2 & 5 & 22.2 & 40.6 & 5 \\
\midrule
\multicolumn{15}{l}{\textit{\textbf{Dense Retrieval}}} \\
E5 & \textbf{55.7} & \textbf{77.3} & 5 & \textbf{61.6} & \textbf{77.8} & 5 & 32.3 & 52.0 & 5 & 51.7 & \textbf{70.9} & 5 & 21.6 & 39.8 & 5 \\
BGE & 50.3 & 73.6 & 5 & 58.7 & 75.1 & 5 & 33.7 & 54.7 & 5 & 50.8 & 69.6 & 5 & 21.1 & 38.9 & 5 \\
\midrule
\multicolumn{15}{l}{\textit{\textbf{Generative Retrieval}}} \\
Naive Constrain & 13.1 & 26.9 & 5 & 23.0 & 46.9 & 5 & 11.8 & 21.6 & 5 & 10.9 & 21.2 & 5 & 9.4 & 19.0 & 5 \\
\rowcolor[RGB]{236,244,252} 
RetroLLM & 51.6 & 62.5 & \textbf{3.20} & 61.1 & 71.0 & \textbf{2.80} & \textbf{35.6} & \textbf{57.3} & \textbf{3.86} & \textbf{57.0} & 70.1 & \textbf{4.07} & \textbf{23.0} & \textbf{41.8} & \textbf{4.40} \\
\bottomrule
\end{tabular}
\end{table*}

\begin{table*}[ht]
\centering
\caption{Detailed performance comparison of RetroLLM using various base models, including the Llama3 series, Qwen-2.5 series, and Mistral series, with parameter sizes ranging from 1B to 14B. All base models we used are the instruction-tuned versions. The best results are highlighted in \textbf{Bold}.}
\label{tab:base_model_app}
\setlength\tabcolsep{5pt}
\fontsize{8.9pt}{10.6pt}\selectfont
\begin{tabular}{p{2.4cm}ccccccccccccccc}
\toprule
 & \multicolumn{9}{c}{\textbf{In-domain Datasets}} & \multicolumn{6}{c}{\textbf{Out-of-domain Datasets}} \\
\cmidrule(lr){2-10} \cmidrule(lr){11-16}
\textbf{Base Model} & \multicolumn{3}{c}{\textbf{NQ}} & \multicolumn{3}{c}{\textbf{TriviaQA}} & \multicolumn{3}{c}{\textbf{HotpotQA}} & \multicolumn{3}{c}{\textbf{PopQA}} & \multicolumn{3}{c}{\textbf{2WIKI}} \\
\cmidrule(lr){2-4} \cmidrule(lr){5-7} \cmidrule(lr){8-10} \cmidrule(lr){11-13} \cmidrule(lr){14-16}
 & Acc & F1 & Tok & Acc & F1 & Tok & Acc & F1 & Tok & Acc & F1 & Tok & Acc & F1 & Tok \\
\midrule
\multicolumn{15}{l}{\textit{\textbf{Llama3 Series}}} \\
Llama3.2-1B & 54.4 & 35.8 & 260 & 64.4 & 52.9 & 288 & 58.8 & 33.5 & 573 & 63.3 & 32.9 & 344 & 44.5 & 28.5 & \textbf{583} \\
Llama3.2-3B & 58.9 & 45.4 & 278 & 67.8 & 62.1 & 267 & 61.3 & 37.8 & 609 & 64.7 & 40.4 & 338 & 47.3 & 32.2 & 632 \\
Llama3-8B   & 59.2 & 46.4 & 306 & 72.7 & 69.3 & 256 & 62.2 & \textbf{47.4} & 575 & 65.2 & 41.4 & 338 & 48.7 & 36.1 & 668 \\
\midrule
\multicolumn{15}{l}{\textit{\textbf{Qwen2.5 Series}}} \\
Qwen2.5-1.5B & 50.1 & 34.3 & \textbf{200} & 57.2 & 51.2 & \textbf{170} & 57.0 & 32.6 & \textbf{539} & 59.5 & 32.6 & \textbf{286} & 47.5 & 26.3 & 650 \\
Qwen2.5-3B   & 52.1 & 36.8 & 236 & 61.4 & 56.3 & 212 & 60.6 & 34.1 & 628 & 64.0 & 34.8 & 336 & 48.1 & 30.6 & 694 \\
Qwen2.5-7B   & 54.9 & 42.3 & 230 & 64.5 & 62.4 & 196 & 61.9 & 42.0 & 549 & 62.8 & 37.1 & 313 & 48.7 & 32.5 & 634 \\
Qwen2.5-14B  & 58.6 & \textbf{50.6} & 225 & 72.8 & 69.5 & 186 & \textbf{62.6} & 45.9 & 568 & 64.3 & 40.8 & 343 & \textbf{51.3} & \textbf{36.9} & 687 \\
\midrule
\multicolumn{15}{l}{\textit{\textbf{Mistral Series}}} \\
\rowcolor[RGB]{236,244,252} 
Mistral-7B   & \textbf{61.6} & 49.8 & 302 & \textbf{74.3} & \textbf{72.8} & 287 & 61.9 & 47.2 & 607 & \textbf{65.7} & \textbf{43.0} & 355 & 48.9 & 36.2 & 661 \\
\bottomrule
\end{tabular}
\end{table*}

\subsubsection{Training}
\label{app:training}

\paragraph{Training Data Construction}
The data construction approach simulates the model's inference process. For each labeled QA pair $(q,a)$, we first utilize a sparse retriever SPLADE-v3~\cite{splade-v3} to obtain top-8 clues $\mathcal{C}_\text{exp}$ and retrieve top-20 documents. We then locate the sentences containing these clues within the documents, followed by employing a reranker to obtain the top-$k_e$ relevant evidences $\mathcal{E}_\text{rel}$, where $k_e$ is set to 5 for single-hop QA and 10 for multi-hop QA tasks. Next, we examine whether the labeled answer $a$ is contained within each evidence $e$ to determine if the evidence can address the original query $q$. To further ensure labeling accuracy, we employ a \texttt{Llama3.1-70B-Instruct}~\cite{llama3} model to judge whether each $e \in \mathcal{E}_\text{rel}$ can genuinely answer the query $q$. We consider an evidence $e$ relevant only if it both contains $a$ and is labeled as relevant by the LLM. Subsequently, we select the top-$k \leq k_e$ evidences where the $k$-th evidence is the first relevant $e$. For target clues, we utilize \texttt{Llama3.1-70B-Instruct} to extract key entities from the query and relevant evidences to construct target clues $\mathcal{C}_\text{gen}$. This process yields the training pair $(q,\mathcal{C}_\text{gen},\mathcal{E},a)$, with the target format illustrated in Figure~\ref{fig:overview} and Appendix~\ref{app:case_study}.

\paragraph{Model Optimization}
As described in Section~\ref{sec:training}, we use a standard sequence-to-sequence loss to train the model. For efficient model training, we employ LoRA~\cite{lora}, setting \texttt{lora\_r} to 16 and \texttt{lora\_alpha} to 64. We set training epochs to 3 and $\gamma$ to 2. Since evidence generation is performed under constraints, most of the middle tokens in evidence generation have limited choices under the constraints of the FM-Index; the crucial parts are the first few tokens at the beginning of the evidence and the tokens at the end that decide to finish the evidence. Therefore, we set the middle 80\% tokens of each evidence not to participate in training, so that the model training focuses more on the key parts. Since we added special tokens to represent the start, separation, and end operations of clue and evidence generation, in the model parameters trained, besides the parameters trained by LoRA, we also added the embeddings corresponding to the new tokens to effectively learn the generation of new tokens.

\subsubsection{Inference}
As illustrated in Figure~\ref{fig:overview}, RetroLLM includes the following three stages. In the clue generation stage, RetroLLM first generates clues with corpus-level FM-Index constraints. The format of this part is ``\texttt{<|clue|>} $c_1$ \texttt{<|sep|>} $c_2$ \texttt{<|sep|>} ... \texttt{<|/clue|>}''. During clue generation, we simultaneously expand clues with the sparse lexical and expansion model SPLADE-v3~\cite{splade-v3}. We set the number of expanded clues to 8 and the maximum number of generated clues to 5.

In the evidence generation stage, evidence is generated based on document-level FM-Index constraints and future window relevance. The format of this part is ``\texttt{<|evidence|>} $e_1$ \texttt{<|sep|>} $e_2$ \texttt{<|sep|>} ... \texttt{<|/evidence|>}''. We limit the maximum number of generated evidence for single-hop and multi-hop QA to 5 and 10, respectively. We set $w_1$ and $w_2$ to 1 and 2, respectively, and $\lambda$ to 100.

In the answer generation stage, no constraints are added during the final answer generation.

\begin{table*}[ht]
\centering
\caption{Detailed performance with different number of generated evidence.}
\label{tab:evidence_num_app}
\setlength\tabcolsep{8pt}
\fontsize{8.9pt}{11pt}\selectfont
\begin{tabular}{p{1.5cm}<{\centering}cccccccccccccc}
\toprule
 & \multicolumn{6}{c}{\textbf{In-domain Datasets}} & \multicolumn{4}{c}{\textbf{Out-of-domain Datasets}} \\
\cmidrule(lr){2-6} \cmidrule(lr){7-11}
\textbf{\# Num} & \multicolumn{2}{c}{\textbf{NQ}} & \multicolumn{2}{c}{\textbf{TriviaQA}} & \multicolumn{2}{c}{\textbf{HotpotQA}} & \multicolumn{2}{c}{\textbf{PopQA}} & \multicolumn{2}{c}{\textbf{2WIKI}} \\
\cmidrule(lr){2-3} \cmidrule(lr){4-5} \cmidrule(lr){6-7} \cmidrule(lr){8-9} \cmidrule(lr){10-11}
 & Acc & F1 & Acc & F1 & Acc & F1 & Acc & F1 & Acc & F1 \\
\midrule
\textbf{1}  & 42.2 & 40.5 & 59.3 & 61.6 & 50.6 & 44.2 & 43.9 & 40.9 & 35.1 & 31.3 \\
\textbf{2}  & 50.6 & 42.3 & 66.3 & 65.9 & 59.8 & 43.8 & 52.8 & 45.9 & 39.8 & 34.6 \\
\textbf{3}  & 54.4 & 42.5 & 69.3 & 67.2 & 61.9 & 43.0 & 55.7 & 45.5 & 42.1 & 34.5 \\
\textbf{4}  & 56.7 & 43.1 & 70.9 & 67.6 & 64.6 & 41.0 & 57.7 & 45.7 & 43.9 & 34.8 \\
\textbf{5}  & 61.5 & 49.4 & 74.6 & 72.9 & 66.8 & 43.0 & 59.4 & 46.8 & 45.9 & 36.6 \\
\textbf{6}  & 61.7 & 49.5 & 74.6 & 73.0 & 67.4 & 42.8 & 60.1 & 47.1 & 47.9 & 37.1 \\
\textbf{7}  & 61.7 & 49.5 & 74.6 & 72.9 & 67.6 & 42.5 & 60.8 & 47.0 & 48.4 & 36.5 \\
\textbf{8}  & 61.7 & 49.5 & 74.6 & 72.9 & 68.0 & 42.7 & 61.2 & 46.9 & 48.6 & 37.2 \\
\textbf{9}  & 61.7 & 49.5 & 74.6 & 72.9 & 68.0 & 42.7 & 61.6 & 47.1 & 48.7 & 37.0 \\
\textbf{10} & 61.7 & 49.5 & 74.6 & 72.9 & 68.5 & 42.7 & 61.9 & 47.1 & 48.9 & 36.2 \\
\bottomrule
\end{tabular}
\end{table*}

\section{Detailed Experimental Results}
\label{app:detailed_results}
This section presents detailed experimental results and analysis, including retrieval performance, the impact of RetroLLM performance with different base models and generated evidence quantity.

\subsection{Analysis of Retrieval Performance}
We analyze the retrieval performance of RetroLLM compared to sparse and dense retrieval baselines, as discussed in Section~\ref{par:retrieval_analysis}. The results are shown in Table~\ref{tab:retrieval_performance_app}

(1) For single-hop QA tasks, RetroLLM demonstrates superior accuracy on R@1, thanks to the design of clues and future windows, which help precisely locate the relevant evidence. For instance, on the PopQA dataset, RetroLLM achieves an R@1 of \textbf{57.0\%}, surpassing the best dense retriever E5, which attains \textbf{51.7\%}. Additionally, RetroLLM uses fewer passages on average (\textbf{2.80} for TriviaQA and \textbf{3.20} for NQ) compared to the fixed number of 5 in baseline methods, indicating more efficient retrieval.

(2) For multi-hop QA tasks, RetroLLM shows superior accuracy compared to all other methods for both R@1 and R@5, while utilizing a smaller average number of retrieved passages. Specifically, on HotpotQA, RetroLLM achieves an R@1 of \textbf{35.6\%}, outperforming E5's \textbf{32.3\%} and SPLADE-v3's \textbf{32.9\%}. On the 2WIKI dataset, RetroLLM attains an R@1 of \textbf{23.0\%}, higher than E5's \textbf{21.6\%}, demonstrating its effectiveness in multi-hop retrieval scenarios while using only \textbf{4.40} passages on average versus the baseline's 5.

(3) Notably, the naive generative retrieval method using constrained beam search performs poorly on all metrics, further validating the severity of false pruning, as discussed in Section~\ref{sec:empirical_study}. For example, on the NQ dataset, the naive method achieves an R@1 of only \textbf{13.1\%}, significantly lower than RetroLLM's \textbf{51.6\%}. Similarly, on TriviaQA, it attains an R@1 of \textbf{23.0\%} compared to RetroLLM's \textbf{61.1\%}, highlighting the substantial performance gap and the advantages of our approach.

\subsection{Impact of Different Base Models}
\label{app:detailed_results_base_llm}
To evaluate the performance of RetroLLM using different backbone LLMs with varying parameter sizes, we conducted experiments using the Mistral, Llama3, and Qwen2.5 series, with parameters ranging from 1B to 14B, as discussed in Section~\ref{par:base_model_analysis}. The results are shown in Figure~\ref{tab:base_model_app}. We observe that:

(1) \textbf{As the parameter size increases, RetroLLM's performance steadily improves, aligning with the scaling law.} For example, within the Llama3 series, the accuracy on the NQ dataset rises from 54.4\% for Llama3.2-1B to 59.2\% for Llama3-8B. Similarly, the F1 score on TriviaQA improves from 52.9\% to 69.3\%. In the Qwen2.5 series, the accuracy on NQ increases from 50.1\% for Qwen2.5-1.5B to 58.6\% for Qwen2.5-14B, and the F1 score climbs from 34.3\% to 50.6\%. This consistent enhancement across different model sizes indicates that larger base models contribute to better retrieval performance in RetroLLM.

(2) \textbf{There are slight performance differences across the different models (Mistral, Llama3, Qwen2.5), with Mistral generally outperforming Llama3, which in turn outperforms Qwen2.5.} Specifically, Mistral-7B achieves the highest accuracy on several datasets, such as \textbf{61.6\%} on NQ and \textbf{74.3\%} on TriviaQA, surpassing both Llama3-8B and Qwen2.5-14B. On the PopQA dataset, Mistral-7B attains an accuracy of \textbf{65.7\%}, compared to 65.2\% for Llama3-8B and 64.3\% for Qwen2.5-14B. Despite these variations, all models confirm the effectiveness of RetroLLM, as even smaller models like Qwen2.5-1.5B achieve reasonable performance (e.g., 50.1\% accuracy on NQ and 57.2\% on TriviaQA), demonstrating that RetroLLM is robust across different base models and parameter sizes.

\subsection{Impact of Generated Evidence Quantity}
Since RetroLLM can dynamically determine the number of evidence to retrieve, we investigated the effect of different maximum retrieval quantities on performance, as discussed in Section~\ref{par:evidence_num_analysis}. The results are shown in Table~\ref{tab:evidence_num_app}. We observe that:

(1) \textbf{When retrieving up to 1-5 evidence, performance continues to improve as the number of retrieved pieces increases, suggesting that more evidence contributes to stronger performance in these tasks.} For instance, on the NQ dataset, the accuracy improves from 42.2\% when retrieving only one piece of evidence to 61.5\% with five pieces. Similarly, the accuracy on TriviaQA rises from 59.3\% to 74.6\% as the number increases from one to five. This trend indicates that accessing more evidence enables RetroLLM to retrieve relevant information more effectively, enhancing answer accuracy.

(2) \textbf{However, for multi-hop QA, performance stabilizes around 6 evidence, as more evidence can bring in both useful and distracting information, thereby limiting further performance gains.} Specifically, on the HotpotQA dataset, the accuracy increases from 50.6\% with one piece of evidence to 67.4\% with six pieces, but additional evidence beyond this point yields diminishing returns (e.g., 68.5\% accuracy at ten pieces). This suggests that while some additional evidence is beneficial, too much can introduce noise that counteracts the benefits, highlighting the importance of a balanced retrieval strategy.

\section{Case Study}
\label{app:case_study}
This section presents examples from RetroLLM and compares them with outputs from a naive constrained beam search method. These examples illustrate the detailed workings of our method and highlight the shortcomings of the naive approach.

\subsection{Examples from RetroLLM}
Tables~\ref{tab:case_single} and \ref{tab:case_multi} display examples from single-hop and multi-hop question-answering (QA) datasets, respectively. The overall process of RetroLLM consists of two main stages: \emph{clue generation} and \emph{evidence generation}.

In the \textbf{clue generation} stage, RetroLLM identifies key terms or phrases related to the question, encapsulated within the \red{<|clue|>} and \red{<|/clue|>} tokens, separated by \red{<|sep|>} token. These clues serve to focus the model's attention on pertinent concepts or entities that are crucial for answering the question. For instance, in Example \#1 from the NQ Dataset (Table~\ref{tab:case_single}), the clue generated is ``The Star,'' which directly relates to the movie mentioned in the question.

Following clue generation, the \textbf{evidence generation} stage involves retrieving relevant information from the knowledge corpus, guided by the previously identified clues. The retrieved evidence is enclosed within the \blue{<|evidence|>} and \blue{<|/evidence|>} tokens. This evidence often comprises multiple snippets of information, separated by \blue{<|sep|>} token, which collectively support the final answer. For example, in Example \#1 from the HotpotQA Dataset (Table~\ref{tab:case_multi}), the model generates evidence about ``Old and in the Way'' and ``Owsley Stanley,'' providing a coherent chain of information that leads to the correct answer.

\subsection{Comparing RetroLLM with Naive Beam Search Method}
Table~\ref{tab:case_compare_nq} and \ref{tab:case_compare_tqa} compares the outputs of the naive constrained beam search method with those of RetroLLM for a question from the NQ Dataset. The naive method attempts to generate evidence under corpus-level FM-Index constraints, but this approach leads to several issues.

The beams generated by the naive method contain evidence that is largely irrelevant or incoherent. Although some initial phrases may appear related to the question, the continuation often deviates significantly, producing sentences that do not contribute to answering the question correctly. For instance, the naive method incorrectly identifies ``Roger Maris,'' ``1903,'' and ``Jonathan Elliot'' as answers to the question ``who got the first nobel prize in physics?'' These incorrect answers result from the model's inability to maintain topic coherence under the stringent corpus-level constraints, a phenomenon known as the \emph{false pruning problem} discussed in Section~\ref{sec:empirical_study} and Appendix~\ref{app:false_pruning}.

In contrast, RetroLLM addresses this issue by first generating clues to narrow down the relevant document subset and then performing forward-looking constrained evidence generation within this subset. This method reduces false pruning and enhances the accuracy of evidence retrieval. 

\begin{table*}[!tbp]
    \centering
    \caption{Examples from RetroLLM on single-hop QA datasets, with special tokens used in the clue and evidence generation stages highlighted in \red{orange box} and \blue{blue box}, respectively. All generated evidence is from Wikipedia.}
    \fontsize{9pt}{11pt}\selectfont
    \begin{tabular}{p{0.98\linewidth}}
    \midrule
        \rowcolor{gray!20}\textbf{Example \#1 from NQ Dataset} \\
    \midrule
        \textbf{Question:} \\
        when does the movie the star come out? \\
        \textbf{Labeled Answer:} \\
        \textcolor{mygreen}{["November 17, 2017"]} \\
    \midrule
        \textbf{Results by RetroLLM}\\
    \midrule
        \textbf{Model Input:} \\
        Question: when does the movie the star come out?\\
        Your Response: \\
    \midrule
        \textbf{Model Output:} \\
         \red{<|clue|>} The Star \red{<|/clue|>} \blue{<|evidence|>} the film, then titled "The Lamb", was given an official release date of December 8, 2017. On June 20, 2016, it was announced that Brian Henson and Lisa Henson from the Jim Henson Company would be executive producers for the film, now titled "The Star", making it the first Henson-produced film for Sony since "The Adventures of Elmo in Grouchland" in 1999. \blue{<|sep|>} The film's title was changed from the working title "Dog Years", in a mutual decision between the filmmakers and A24 (who made the suggestion). It was released on February 22, 2018, through DirecTV Cinema before being released in a limited release and through video on demand on March 30, 2018. \blue{<|sep|>} Chenoweth, Tracy Morgan, Tyler Perry, and Oprah Winfrey. The film was released by Columbia Pictures through its parental label Sony Pictures Entertainment on \textcolor{mygreen}{November 17, 2017} in the United States. The film received mixed reviews and grossed over \$62 million worldwide. \blue{<|/evidence|>} The answer is \textcolor{mygreen}{November 17, 2017}. \\
         
    \midrule
        \rowcolor{gray!20}\textbf{Example \#2 from TriviaQA Dataset} \\
    \midrule
        \textbf{Question:} \\
        Who was the man behind The Chipmunks? \\
        \textbf{Labeled Answer:} \\
        \textcolor{mygreen}{["David Seville"]} \\
    \midrule
        \textbf{Results by RetroLLM}\\
    \midrule
        \textbf{Model Input:} \\
        Question: Who was the man behind The Chipmunks?\\
        Your Response: \\
    \midrule
        \textbf{Model Output:} \\
        \red{<|clue|>} The Chipmunks \red{<|/clue|>} \blue{<|evidence|>} 1972), better known by the stage name \textcolor{mygreen}{David Seville}, was an American singer-songwriter and actor, best known for creating the novelty musical group The Chipmunks (now better known as "Alvin and the Chipmunks"). Initially a stage and film actor, he rose to prominence in 1958 with the songs "Witch Doctor" and "The Chipmunk Song (Christmas Don't Be Late)", which both became "Billboard" number-one singles. \blue{<|/evidence|>} The answer is \textcolor{mygreen}{David Seville}. \\

    \midrule
        \rowcolor{gray!20}\textbf{Example \#3 from PopQA Dataset} \\
    \midrule
        \textbf{Question:} \\
        What is Carsten Carlsen's occupation? \\
        \textbf{Labeled Answer:} \\
        \textcolor{mygreen}{["pianist", "composer"]} \\
    \midrule
        \textbf{Results by RetroLLM}\\
    \midrule
        \textbf{Model Input:} \\
        Question: What is Carsten Carlsen's occupation?\\
        Your Response: \\
    \midrule
        \textbf{Model Output:} \\
        \red{<|clue|>} Carlsen \red{<|/clue|>} \blue{<|evidence|>} Carsten Carlsen (5 June 1892 – 28 August 1961) was a Norwegian \textcolor{mygreen}{pianist} and \textcolor{mygreen}{composer}. Carsten Marensius Carlsen was born in Oslo, Norway. His parents were Anton Carlsen (1862–1943) and Louise Larsen (1876–1957). \blue{<|sep|>} Carlsen has directed more than 25 films, 19 for which he wrote the screenplay. In 2006, he received the Golden Swan Lifetime Achievement Award at the Copenhagen International Film Festival. Carlsen was born on 4 June 1927 in Aalborg, Denmark. \blue{<|/evidence|>} The answer is \textcolor{mygreen}{pianist} and \textcolor{mygreen}{composer}. \\

    \bottomrule
    \end{tabular}
    \label{tab:case_single}
\end{table*}

\begin{table*}[!tbp]
    \centering
    \caption{Examples from RetroLLM on multi-hop QA datasets, with special tokens used in the clue and evidence generation stages highlighted in \red{orange box} and \blue{blue box}, respectively. All generated evidence is from Wikipedia.}
    \fontsize{9pt}{11pt}\selectfont
    \begin{tabular}{p{0.98\linewidth}}
    \midrule
        \rowcolor{gray!20}\textbf{Example \#1 from HotpotQA Dataset} \\
    \midrule
        \textbf{Question:} \\
        Which American audio engineer and clandestine chemist, who was a key figure in the San Francisco Bay Area hippie movement during the 1960s recorded the album "Old and in the Way?"\\
        \textbf{Labeled Answer:} \\
        \textcolor{mygreen}{["Owsley Stanley"]} \\
    \midrule
        \textbf{Results by RetroLLM}\\
    \midrule
        \textbf{Model Input:} \\
        Question: Which American audio engineer and clandestine chemist, who was a key figure in the San Francisco Bay Area hippie movement during the 1960s recorded the album "Old and in the Way?"\\
        Your Response: \\
    \midrule
        \textbf{Model Output:} \\
        \red{<|clue|>} Old and in the Way \red{<|sep|>} Owsley Stanley \red{<|/clue|>} \blue{<|evidence|>} \textcolor{mygreen}{Owsley Stanley} III (January 19, 1935 – March 12, 2011) was an American audio engineer and clandestine chemist. He was a key figure in the San Francisco Bay Area hippie movement during the 1960s and played a pivotal role in the decade’s counterculture. \blue{<|sep|>} Old \& In the Way (album) Old \& In the Way is the self-titled first album by the bluegrass band Old \& In the Way. It was recorded 8 October 1973 at The Boarding House in San Francisco by \textcolor{mygreen}{Owsley Stanley} and Vickie Babcock. \blue{<|sep|>} The 1960s, elements of the expanding Beat movement were incorporated into the hippie and larger counterculture movements. Neal Cassady, as the driver for Ken Kesey's bus Furthur, was the primary bridge between these two generations. Ginsberg's work also became an integral element of early 1960s hippie culture. \blue{<|sep|>} In San Francisco, including Quicksilver Messenger Service, Jefferson Airplane, early Jefferson Starship, Old \& In the Way, Janis Joplin, Big Brother and the Holding Company, Taj Mahal, Santana, Miles Davis, The Flying Burrito Brothers, Jimi Hendrix, Johnny Cash, and Blue Cheer. \blue{<|/evidence|>} The answer is \textcolor{mygreen}{Owsley Stanley}. \\
         
    \midrule
        \rowcolor{gray!20}\textbf{Example \#2 from 2WikiMultiHopQA Dataset} \\
    \midrule
        \textbf{Question:} \\
        Where was the director of film Ronnie Rocket born? \\
        \textbf{Labeled Answer:} \\
        \textcolor{mygreen}{["Missoula", "Missoula, Montana"]} \\
    \midrule
        \textbf{Results by RetroLLM}\\
    \midrule
        \textbf{Model Input:} \\
        Question: Where was the director of film Ronnie Rocket born?\\
        Your Response: \\
    \midrule
        \textbf{Model Output:} \\
        \red{<|clue|>} Ronnie Rocket \red{<|/clue|>} \blue{<|evidence|>} Ronnie Rocket is an unfinished film project written by David Lynch, who also intended to direct it. Begun after the success of Lynch's 1977 film "Eraserhead", "Ronnie Rocket" was shelved after Lynch felt he would be unable to find financial backing for the project. \blue{<|sep|>} The script to the AFI, but they rejected it as they felt the planned 45-minute runtime was too long for such a figurative, nonlinear script. After the success of "Eraserhead", Lynch decided to focus on the screenplay of his second film, entitled "Ronnie Rocket", which he also wanted to direct. \blue{<|sep|>} He can use to produce music or cause destruction. The boy names himself Ronnie Rocket and becomes a rock star, befriending a tap-dancer named Electra-Cute. The film was to make use of several themes that have since become recurring elements within David Lynch's works, with a write-up for "The A.V. \blue{<|sep|>} Dave Dennison, Max and Dani Dennison's father, in Disney's "Hocus Pocus". Rocket was born in Bangor, Maine, the son of Mary Aurelia (Fogler) and Sumner Abbott "Ham" Claverie. \blue{<|sep|>} In 2008, Refn co-founded the Copenhagen-based production company Space Rocket Nation. Refn was born in Copenhagen, Denmark, and raised partly in New York, United States. Refn's parents are Danish film director and editor Anders Refn and cinematographer Vibeke Winding. \blue{<|sep|>} The two realized that "Ronnie Rocket" was unlikely to find sufficient financing to be produced, Lynch asked to see some already-written scripts to work from for his next film instead. Cornfeld found four scripts he felt Lynch would be interested in, but on hearing the name of the first, the director decided his next project would be "The Elephant Man". \blue{<|sep|>} His next film, settling on what would become 1980's "The Elephant Man". "Ronnie Rocket" was to feature many of the elements which have since come to be seen as Lynch's hallmarks; including industrial art direction, 1950s popular culture and physical deformity. \blue{<|sep|>} He shared with Billy Williams. Taylor was born in Hampstead, London and entered the film industry in his late teens working for Gainsborough Pictures at Lime Grove in Shepherds Bush. Taylor's first film was as a clapper boy on "The Young Mr Pitt" (1942). \blue{<|sep|>} In his films have earned them a reputation as works that "disturb, offend or mystify" general audiences. Although born in \textcolor{mygreen}{Missoula, Montana}, Lynch spent his youth traveling across the United States due to his father Donald's job for the Department of Agriculture; as a result, Lynch attended school across several states. \blue{<|/evidence|>} The answer is \textcolor{mygreen}{Missoula, Montana}. \\

    \bottomrule
    \end{tabular}
    \label{tab:case_multi}
\end{table*}

\begin{table*}[!tbp]
    \centering
    \caption{An example comparing outputs from RetroLLM and the naive constrained beam search method. Special tokens used during the clue and evidence generation stages are highlighted in \red{orange boxes} and \blue{blue boxes}, respectively. Content colored in \textcolor{mygreen}{green} indicates correct (or partially correct) answers, whereas content colored in \textcolor{red}{red} indicates incorrect answers. All generated evidence is from Wikipedia.}
    \fontsize{9pt}{11pt}\selectfont
    \begin{tabular}{p{0.98\linewidth}}
    \midrule
        \rowcolor{gray!20}\textbf{Example \#1 from NQ Dataset} \\
    \midrule
        \textbf{Question:} \\
        who got the first nobel prize in physics? \\
        \textbf{Labeled Answer:} \\
        \textcolor{mygreen}{["Wilhelm Conrad Röntgen"]} \\

    \midrule
        \rowcolor{gray!20}
        \textbf{Results by RetroLLM}\\
    \midrule
        \textbf{Model Input:} \\
        Question: who got the first nobel prize in physics?\\
        Your Response: \\
    \midrule
        \textbf{Model Output:} \\
        \red{<|clue|>} Nobel Prize \red{<|sep|>} Physics \red{<|/clue|>} \blue{<|evidence|>} the others being the Nobel Prize in Chemistry, Nobel Prize in Literature, Nobel Peace Prize, and Nobel Prize in Physiology or Medicine. The first Nobel Prize in Physics was awarded to physicist \textcolor{mygreen}{Wilhelm Röntgen} in recognition of the extraordinary services he rendered by the discovery of the remarkable rays (or X-rays). \blue{<|/evidence|>} The answer is \textcolor{mygreen}{Wilhelm Röntgen}.\\

    \midrule
        \rowcolor{gray!20}
        \textbf{Results by Naive Constrained Beam Search}\\
    \midrule
        \textbf{Model Input:} \\
        Question: who got the first nobel prize in physics?\\
        Your Response: \\
    \midrule
        \textbf{Model Output (beam\_size = 5):} \\
        \textbf{Beam 1:} \blue{<|evidence|>} in 1901, the first year of play for the American League. The modern-era record for lowest batting average for a player that qualified for the batting title is .197 by roger maris in 1962. \blue{<|/evidence|>} The answer is \textcolor{red}{Roger Maris}.\\
        \textbf{Beam 2:} \blue{<|evidence|>} the first physics and chemistry laboratory in a Danish school in 1903, the inclusion of housekeeping in the curriculum, and the first girls' school to include carpentry in the curriculum in 1904. \blue{<|/evidence|>} The answer is \textcolor{red}{1903}.\\
        \textbf{Beam 3:} \blue{<|evidence|>} william johnson Jonathan Elliot (historian) Jonathan Elliot (1784\u20131846) was a 19th-century \blue{<|/evidence|>} The answer is \textcolor{red}{Jonathan Elliot}.\\
        \textbf{Beam 4:} \blue{<|evidence|>} the first physics Nobel prize winner born in an Arab country. In 2015, Cohen-Tannoudji signed the \"Mainau Declaration 2015\" which calls for a boycott of israel. \blue{<|/evidence|>} The answer is \textcolor{red}{Cohen - Tannoudji}.\\
        \textbf{Beam 5:} \blue{<|evidence|>} the first prize was awarded in 2004. He was the sixth of the seven children in the Shaw family, and was nicknamed Uncle Six (\"Luk Sook\"). \blue{<|/evidence|>} The answer is \textcolor{red}{He was the sixth of the seven children in the Shaw family}. \\

    \bottomrule
    \end{tabular}
    \label{tab:case_compare_nq}
\end{table*}

\begin{table*}[!tbp]
    \centering
    \caption{An example comparing outputs from RetroLLM and the naive constrained beam search method. Special tokens used during the clue and evidence generation stages are highlighted in \red{orange boxes} and \blue{blue boxes}, respectively. Content colored in \textcolor{mygreen}{green} indicates correct answers, whereas content colored in \textcolor{red}{red} indicates incorrect answers. All generated evidence is from Wikipedia.}
    \fontsize{9pt}{11pt}\selectfont
    \begin{tabular}{p{0.98\linewidth}}
    \midrule
        \rowcolor{gray!20}\textbf{Example \#2 from TriviaQA Dataset} \\
    \midrule
        \textbf{Question:} \\
        Who was the man behind The Chipmunks? \\
        \textbf{Labeled Answer:} \\
        \textcolor{mygreen}{["David Seville"]} \\

    \midrule
        \rowcolor{gray!20}
        \textbf{Results by RetroLLM}\\
    \midrule
        \textbf{Model Input:} \\
        Question: Who was the man behind The Chipmunks?\\
        Your Response: \\
    \midrule
        \textbf{Model Output:} \\
        \red{<|clue|>} The Chipmunks \red{<|/clue|>} \blue{<|evidence|>} known by the stage name \textcolor{mygreen}{David Seville}, was an American singer-songwriter and actor, best known for creating the novelty musical group The Chipmunks (now better known as "Alvin and the Chipmunks"). Initially a stage and film actor, he rose to prominence in 1958 with the songs "Witch Doctor" and "The Chipmunk Song (Christmas Don't Be Late)," which both became "Billboard" number-one singles. \blue{<|/evidence|>} The answer is \textcolor{mygreen}{David Seville}.\\

    \midrule
        \rowcolor{gray!20}
        \textbf{Results by Naive Constrained Beam Search}\\
    \midrule
        \textbf{Model Input:} \\
        Question: Who was the man behind The Chipmunks?\\
        Your Response: \\
    \midrule
        \textbf{Model Output (beam\_size = 5):} \\
        \textbf{Beam 1:} \blue{<|evidence|>} david was optioned to the Triple-A Richmond Braves to start the 1992 season. He made his major league debut for the Braves on \blue{<|/evidence|>} The answer is \textcolor{red}{David}.\\
        \textbf{Beam 2:} \blue{<|evidence|>} the chipmunks' speech was created by speeding up sound clips of normal speech. In a number of the shorts that followed, many of these same sound clips were reused, but the chipmunks' voices were created by using a higher pitch on the original recordings. \blue{<|/evidence|>} The answer is \textcolor{red}{Ross Bagdasarian}.\\
        \textbf{Beam 3:} \blue{<|evidence|>} alvinophilia is a partialism in which a man or woman is attracted to the human navel. According to a study, it is a moderately prevalent fetish among men, and less so among women. \blue{<|/evidence|>} The answer is \textcolor{red}{Alvin}.\\
        \textbf{Beam 4:} \blue{<|evidence|>} the chipmunks are also mostly supplemental prey but are considered more easily caught than tree squirrels, considering that they are more habitual terrestrial foragers. \blue{<|/evidence|>} The answer is \textcolor{red}{Alvin}.\\
        \textbf{Beam 5:} \blue{<|evidence|>} the chipmunks are also mostly supplemental prey but are considered more easily caught than tree squirrels, considering that they are more habitual terrestrial foragers. \blue{<|/evidence|>} The answer is \textcolor{red}{Tree Squirrels}. \\

    \bottomrule
    \end{tabular}
    \label{tab:case_compare_tqa}
\end{table*}

\end{CJK}
\end{document}